\pdfoutput=1

\documentclass[11pt]{article}

\usepackage[final]{acl}

\usepackage{times}
\usepackage{latexsym}

\usepackage[T1]{fontenc}

\usepackage[utf8]{inputenc}

\usepackage{microtype}

\usepackage{inconsolata}

\usepackage{graphicx}
\usepackage[export]{adjustbox}

\usepackage[cjk]{kotex}         

\usepackage{amsmath}            
\usepackage{amssymb}            
\usepackage{mathtools}          
\usepackage{bm}                 

\usepackage{booktabs}           
\usepackage{multirow}           
\usepackage{hhline}             
\usepackage{tabularx}           
\usepackage{colortbl}           
\usepackage{supertabular}          
\usepackage{longtable}
\usepackage{multicol}

\usepackage{xcolor} 

\usepackage{subcaption}         

\usepackage{algorithm}          
\usepackage{algpseudocode}      

\usepackage{setspace}           
\usepackage{xspace}             
\usepackage[normalem]{ulem}     
\usepackage{footnote}

\usepackage{pifont}             

\usepackage[most]{tcolorbox}    

\usepackage[%
    commandnameprefix=always,   
    defaultcolor=purple         
]{changes}

\usepackage[outline]{contour}   
\usepackage{pgfplots}
\usetikzlibrary{patterns}
\pgfplotsset{compat=1.17}

\definecolor{fontpurple}{RGB}{134, 53, 149}
\definecolor{bgpurple}{RGB}{216,188,214}
\definecolor{metablue}{RGB}{21,144,255}

\newcommand{\fontgray}[1]{\textcolor{gray!80}{\textit{#1}}}
\newcommand{\fontpurple}[1]{\textcolor{fontpurple}{#1}}
\newcommand{\fontblue}[1]{\textcolor{metablue}{#1}}

\useunder{\uline}{\ul}{}

\newcommand{\ie}{\emph{i}.\emph{e}.,\xspace}

\title{
Building Resource-Constrained Language Agents: \\
A Korean Case Study on Chemical Toxicity Information
}

\author{
        \textbf{Hojun Cho}$^\dagger$ \hspace{0.3cm}
        \textbf{Donghu Kim}$^\dagger$ \hspace{0.3cm}
        \textbf{Soyoung Yang}$^\dagger$ \hspace{0.3cm} \\
        \textbf{Chan Lee}$^\S$ \hspace{0.3cm} 
        \textbf{Hunjoo Lee}$^\S$ \hspace{0.3cm}
        \textbf{Jaegul Choo}$^\dagger$ \hspace{0.3cm} \\
        $^\dagger$ KAIST AI \hspace{0.3cm} 
        $^\S$ CHEM. I. NET \\
        \texttt{hojun.cho@kaist.ac.kr} \\
  }

\begin{document}
\maketitle

\begin{abstract}
Language agents powered by large language models (LLMs) face significant deployment challenges in resource-constrained environments, particularly for specialized domains and less-common languages. This paper presents \textit{Tox-chat}, a Korean chemical toxicity information agent devised within these limitations. We propose two key innovations: a context-efficient architecture that reduces token consumption through hierarchical section search, and a scenario-based dialogue generation methodology that effectively distills tool-using capabilities from larger models. Experimental evaluations demonstrate that our fine-tuned 8B parameter model substantially outperforms both untuned models and baseline approaches, in terms of DB faithfulness and preference. Our work offers valuable insights for researchers developing domain-specific language agents under practical constraints.

\end{abstract}

\begin{figure}[t]
    \centering
    \includegraphics[width=\linewidth]{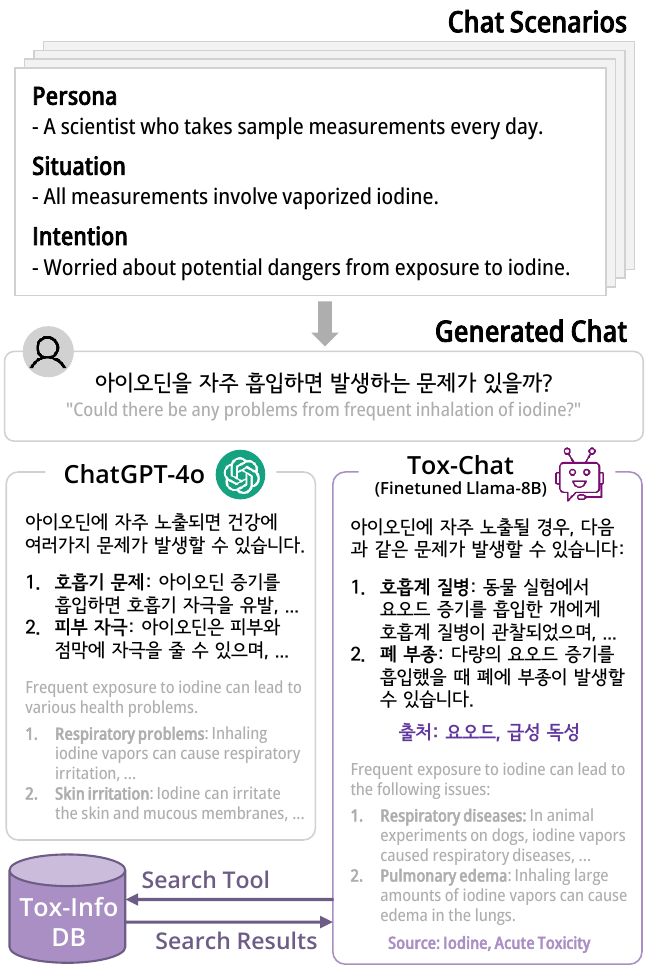}
    \caption{Use case of our model, Tox-chat, in comparison to ChatGPT. Tox-chat can generate grounded answers based on Tox-info, a chemical toxicity database maintained by the Korean government.
    }
    \label{fig:teaser}
    \vspace{-0.5em}
\end{figure}

\section{Introduction}
Language agents are intelligent systems that autonomously perform complex tasks by leveraging various external tools based on large language models (LLMs)~\citep{su-etal-2024-language,xi-etal-2023-rise,wang-etal-2024-survey-large}. 
The core component of a language agent is the LLM that orchestrates the entire system, determining the agent's overall capabilities.

Current state-of-the-art LLMs can be broadly categorized into proprietary models such as ChatGPT~\cite{openai-2024-gpt,openai-2024-gpt4o} and large-scale open-source models like Deepseek-V3~\cite{deepseekai-2025-deepseek}.
However, there are constraints in deploying these models as language agents in practical resource-limited settings, such as government institutions or security-sensitive corporate environments.
Specifically, proprietary models raise concerns in cost and service dependency, while large-scale open-source models demand substantial computational resources.
To address these industrial demands, small open-source LLMs could be employed privately, but they have inherent performance limitations for language agents.
Although it is possible to improve small LLMs capabilities with additional data~\cite{yin-etal-2024-agent,zeng-etal-2024-agenttuning}, dealing with specialized domains or less common languages remain a significant challenge due to data scarcity.

Under these challenging environments, we developed \textit{Tox-chat}, a Korean-based chemical and toxicity information language agent based on a small open-source LLM.
Fig.~\ref{fig:teaser} illustrates a use case of Tox-chat in comparison to ChatGPT.
Tox-chat is a language agent that interacts with Tox-info\footnote{\url{https://www.nifds.go.kr/toxinfo}}, a Wikipedia-style chemical toxicity information database operated by the Korean government.
A detailed description of the Tox-info DB is provided in Appendix~\ref{sec:toxinfo_db_overview}.
When users input questions about chemical exposure or safety in Korean, our model provides grounded and reliable answers based on Tox-info documents.
This system makes complex toxicity information accessible to non-experts, serving as a specialized tool for searching toxicity levels and poisoning data across various chemicals.

In our development of Tox-chat, we aimed to fine-tune LLMs,
which necessitated domain-specific multi-turn tool-using dialogue data.
While creating this dataset, we encountered two significant challenges:
(1) \textbf{Naive retrieval-augmented generation (RAG) is not a viable solution}.
Our initial approach, retrieving relevant information from the database with RAG~\cite{gao-etal-2024-retrieval,xu2024retrieval}, faced limitation in terms of context length and search quality.
Concretely, RAG returns large quantity of text, which would take up a large portion of input context length.
More importantly, these sections were retrieved solely based on text similarity and without verification, making these results unreliable.
(2) \textbf{Training data is scarce and hard to collect}.
Training data for agents that possess both tool-using and multi-turn capabilities is still an emerging research area~\cite{li-etal-2023-api, shim2025tooldial} with limited available datasets.
Moreover, specialized conversational data on Korean chemical toxicity information is virtually non-existent and difficult to collect.

Given these complex constraints, we developed an efficient architecture and training strategy optimized for limited resources.
Our approach introduces two novel technical contributions.

\begin{itemize}
    \item \textbf{Agent structure with hierarchical information retrieval}:
    We constructed a language agent with a hierarchical document-section structure that efficiently searches and summarizes relevant information, while substantially reducing context length requirements.
    This approach extends beyond Wikipedia-style databases to various technical, legal, and medical document repositories with similar organizational structures.

    \item  \textbf{Efficient tool-using dialogue generation}:
    We devised a streamlined methodology for synthesizing multi-turn tool-using dialogue datasets that effectively distills capabilities from proprietary or large-scale LLMs while reflecting user query patterns.
    This enables smaller LLMs to be rapidly adapted as domain-specific language agents with minimal data.

\end{itemize}

We expect that sharing our research experience will provide valuable guidance to researchers developing language agents under practical constraints.

%
%
%

\section{Related Work}
\paragraph{Context-Efficient RAG and Language Agents}
When LLMs generate responses based on RAG, an increased number of retrieved documents not only lengthens the context but also frequently leads to inaccurate outputs due to irrelevant documents~\citep{xu2024retrieval}.
Consequently, there have been attempts to reduce the number of retrieved documents by adaptive selection~\citep{jeong-etal-2024-adaptive,asai2024selfrag}, but still require the agent to process all relevant documents when many are retrieved.
To address these issues, alternative approaches such as RECOMP~\cite{xu2024recomp} and collaborative multi-agent systems~\citep{zhao-etal-2024-longagent,zhang2024chain} attempt to compress the retrieved documents.
On the other hand, there are agentic approaches that directly search for necessary documents to answer queries~\citep{lo-etal-2023-hierarchical,li2024chainofknowledge}.
While these methods avoid context length issues, they may suffer from inefficient inference when retrieval fails.
Therefore, our methodology employs a structure where the language agent first examines summarized relevant documents similar to RECOMP, and then directly searches for and verifies documents as needed.

\paragraph{Dialogue Generation for Agent Fine-Tuning}
There have been numerous attempts to distill the conversational capabilities of state-of-the-art LLMs by generating text data and fine-tuning smaller open-source LLMs~\citep{alpaca,vicuna2023,peng2023instruction,hsieh-etal-2023-distilling}.
Recently, several studies focus on generating tool-using data~\citep{li-etal-2023-api,qin2024toolllm,shim2025tooldial} or aim at distilling agent capabilities~\citep{chen2023fireactlanguageagentfinetuning,yin-etal-2024-agent}.
However, these existing approaches have limitations in fully distilling the tool-using capabilities of modern LLMs, as they typically rely on indirect text generation methods rather than leveraging the actual tool-using mechanisms employed by LLMs.
To overcome these limitations, our approach distills the tool-using capabilities of advanced LLMs via constructing conversations with a user simulator.

\section{Method}
Tox-chat consists of two main components: architecture implementation and constructing dialogue data for distillation.

\subsection{Tox-chat Architecture}
\label{sec:architecture}
\begin{figure}[t]
\begin{center}
\includegraphics[width=\linewidth]{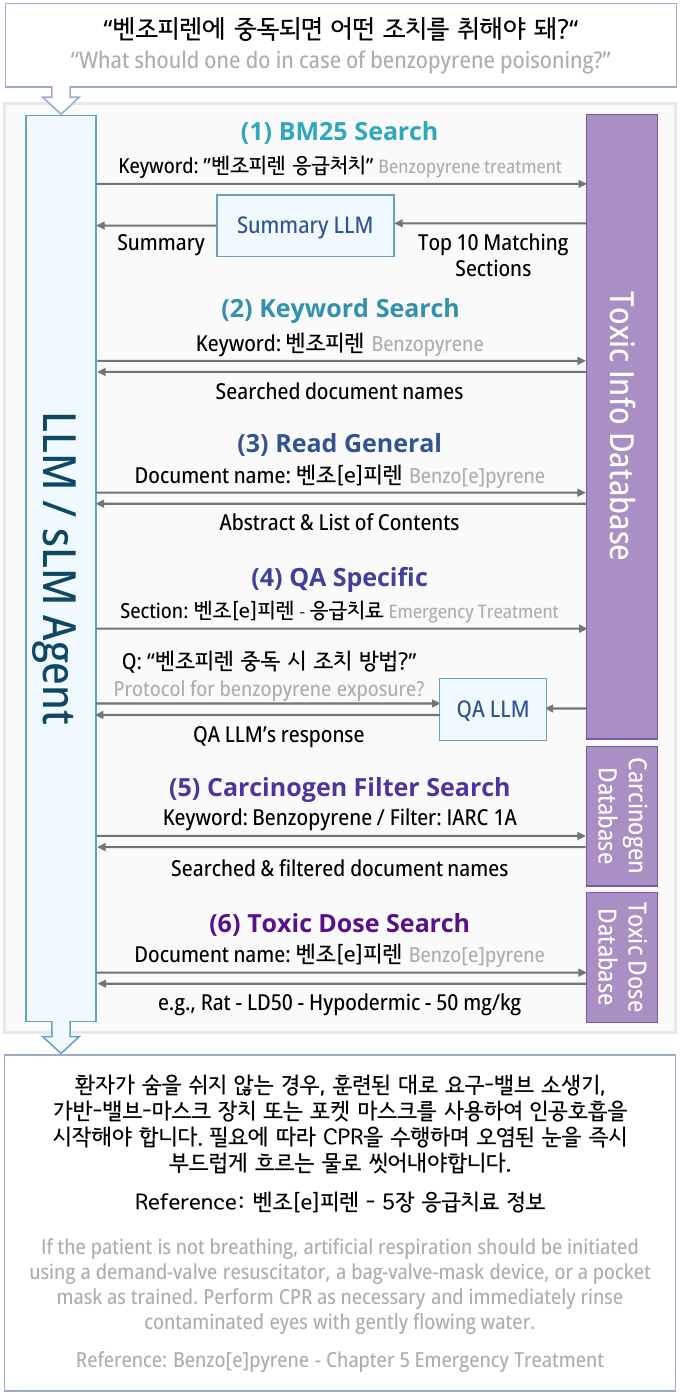}
\end{center}
\vspace{-0.5em}
\caption{Overview of the Tox-chat architecture.}
\label{figure:utd_scaling}
\vspace{-1.0em}
\end{figure}

As shown in Fig.~\ref{figure:utd_scaling}, Tox-chat architecture is the language agent capable of utilizing six tools that enable it to perform searches within the Tox-info database.
All tool-relevant features leverage pre-defined tool-using capabilities of each LLM\footnote{For example, we leverage \href{https://platform.openai.com/docs/guides/function-calling}{function-calling} for ChatGPT and \href{https://www.llama.com/docs/model-cards-and-prompt-formats/llama3_1/\#-tool-calling-(8b/70b/405b)-}{tool-calling} for Llama 3.1.}, and the tool formats also follow the respective settings.

We start with a minimal approach: utilizing a retrieval system.
Each section within the documents is tokenized and indexed as a single chunk using BM25~\citep{robertson-etal-2009-probabilistic}.
By treating each section as a complete unit, we avoid the incomplete retrieval problem \cite{luo-etal-2024-landmark} that occurs when chunks are arbitrarily truncated.
After constructing the BM25 index, we provide the agent with the \texttt{BM25\_Search} tool that retrieves the top-10 document sections most relevant to the search query.
However, including all retrieved sections in the agent's context would significantly increase the context length during multi-turn conversations, leading to substantial increases in computational costs. 
To address this, similar to \citet{xu2024recomp}, we employ a separate summary LLM to deliver condensed results to the agent. 
This summary LLM processes the retrieved sections along with the original query to extract and summarize only the information relevant to the user's request. 
When the retrieved sections lack relevant content, the summary LLM explicitly notifies the agent that no pertinent information is available.

While the \texttt{BM25\_Search} tool provides adequate performance in many scenarios, it struggles when tasked with retrieving specific details about substances. 
In these cases, BM25 often retrieves paragraphs that share lexical similarities with the query but lack substantive relevance, resulting in hallucinations. 
This limitation underscores the critical importance of verifying whether relevant information actually exists within the database.

To address this challenge, we consulted domain experts to understand how humans efficiently search for answers in specialized databases, and incorporated these insights into the tox-chat architecture. We observed that human information-seeking behavior typically follows a structured process:
(1) Search to determine if a relevant document exists.
(2) Open the document to review general information (abstract and table of contents).
(3) Selectively read sections likely to contain the necessary details (e.g., first aid information).

Mimicking this process that provides a more effective framework for retrieving specific substance information than relying solely on BM25 similarity, we designed our main search tool with three components:
(a) \texttt{Keyword\_Search}: Returns a list of documents from the database based on the query.
(b) \texttt{Read\_General}: Returns the Abstract section and table of contents of selected document.
(c) \texttt{QA\_Specific}: Reads a specific section of the selected document.

Similar to \texttt{BM25\_Search}, the detailed section from \texttt{QA\_Specific} can be too long for the agent's context.
To address this, the tool takes an additional "question" input to extract relevant information, and if a section exceeds 200 tokens, a separate QA LLM answers the question and extracts only the necessary details.
Finally, the Tox-info DB provides individual functions for carcinogenicity classification searches and toxicity level searches.
Accordingly, these functions are implemented and provided as the \texttt{Carcinogen\_Filter\_Search} and \texttt{Toxic\_Dose\_Search} tools, respectively.

We instruct the agent to first perform \texttt{BM25\_Searches} before utilizing other search tools, and this helps the agent efficiently search the database.
Nevertheless, the agent maintains autonomy in selecting and using all the tools defined above as needed.
The agent also generates all necessary input arguments for these tools, such as search queries and specific questions, based on the context.

To mitigate hallucination and promote safer responses, we impose a strict \emph{DB-only} constraint through the system prompt, which restricts the agent from generating or reasoning beyond content retrieved from the Tox-info database.
This design helps suppress unsupported or speculative outputs.
For a broader discussion of safety concerns, see the Ethical Considerations section.
Detailed descriptions of each tool and the system prompts used by the agent are provided in Appendix~\ref{sec:prompts}.

\subsection{Distillation to Open-Source LLM}
\label{sec:distillation}


We aim to improve small LLMs to handle tools and generate better responses through supervised fine-tuning (SFT).
For this reason, we need multi-turn tool-using dialogue data based on the Tox-chat architecture.
To generate the training data, we design an LLM-based user simulator to mimic real user interactions. 
We record complete conversation sessions between this simulator and a Tox-chat agent, capturing all aspects including tool usage, search behaviors, and response generation.
This comprehensive dataset, containing the full interaction trace, is then used to fine-tune the small LLM.


\begin{table*}[th]
\centering
\resizebox{\textwidth}{!}{%
\begin{tabular}{cl|cc|cc|c|c}
\toprule
\textbf{Backbone} & \multicolumn{1}{c|}{\textbf{Model}} & \textbf{Fine-Tuned} & \textbf{Tool-Using} & \textbf{\begin{tabular}[c]{@{}c@{}}DB Consist. \\ (\% Success)\end{tabular}} & \textbf{\begin{tabular}[c]{@{}c@{}}Preference\\ (W / L / D)\end{tabular}} & \textbf{\begin{tabular}[c]{@{}c@{}} Search Rate\\ (\% Success)\end{tabular}} & \textbf{\begin{tabular}[c]{@{}c@{}}Response Len.\\ (Avg. Tokens)\end{tabular}}\\
\midrule 
\multirow{4}{*}{GPT} & GPT-4-mini & \textcolor{gray!80}{\ding{55}} & \textcolor{gray!80}{\ding{55}} & 48 & \textbf{80 / 0 / 20} & - & 346.51 \\
 & GPT-4o & \textcolor{gray!80}{\ding{55}} & \textcolor{gray!80}{\ding{55}} & 58 & 78 / 4 / 18 & - & 313.69 \\
 & Tox-chat [GPT-4o-mini] & \textcolor{gray!80}{\ding{55}} & \ding{51} & 80 & 60 / 6 / 34 & \textbf{96} & 219.57 \\
 & Tox-chat [GPT-4o] & \textcolor{gray!80}{\ding{55}} & \ding{51} & \textbf{84} & 52 / 14 / 34 & 94 & 179.45 \\ 
\hline
\multirow{6}{*}{Llama} & Llama-1B & \textcolor{gray!80}{\ding{55}} & \textcolor{gray!80}{\ding{55}} & 14 & 2 / 94 / 4 & - & 336.18 \\
 & Llama-8B & \textcolor{gray!80}{\ding{55}} & \textcolor{gray!80}{\ding{55}} & 6 & - & - & 611.16 \\
 & Tox-chat [Llama-1B] & \textcolor{gray!80}{\ding{55}} & \ding{51} & 4 & 2 / 96 / 2 & 0 & 616.87 \\
 & Tox-chat [Llama-8B] & \textcolor{gray!80}{\ding{55}} & \ding{51} & 44 & 14 / 58 / 28 & 76 & 290.54 \\
 & \textbf{Tox-chat-1B (ours)} & \ding{51} & \ding{51} & 62 & 18 / 42 / 40 & 86 & 189.05 \\
 & \textbf{Tox-chat-8B (ours)} & \ding{51} & \ding{51} & \textbf{68} & \textbf{38 / 34 / 28} & \textbf{92} & 228.22 \\
\bottomrule
\end{tabular}%
}
\caption{Results compare model performance across database consistency (percentage of responses consistent with reference documents), preference metrics (win/lose/draw rates against Llama-8B-Instruct baseline), successful search (percentage of searches that found reference documents), and average response token length. Alternative backbone models are indicated in brackets ({[ ]}). We bold the \textbf{best score} for each backbone model.
}
\label{tab:main-result}
\vspace{-0.5em}
\end{table*}

\paragraph{Scenario Collection}
To effectively simulate users, we first need to clearly define the intended users of Tox-chat. This approach is similar to user modeling techniques~\citep{DBLP:journals/debu/Tan023}. 
To minimize collection costs while ensuring clear representation of user backgrounds and intentions, we define a simplified user data structure called a \textit{scenario}.
Each scenario consists of four key elements:
(a) \textbf{Persona}: The user's personality and characteristics.
(b) \textbf{Situation}: A description of the user's current circumstances.
(c) \textbf{Intention}: The user's purpose for engaging with the agent.
(d) \textbf{Question}: The actual query from the user, representing the first question to the agent.
Collecting data in scenario format rather than isolated questions provides significant advantages for multi-turn data generation. 
When user simulators have clear intentions based on well-defined scenarios, they engage in more meaningful and diverse conversations compared to simply extending interactions from a single initial question. 

We developed these agent usage scenarios in Korean through collaboration with a diverse group including general users, Tox-info specialists, and experts from food safety and pharmaceutical fields. A translated example of such a scenario is shown at the top of Fig~\ref{fig:teaser}. 
While scenario data is easier to create than full dialogues, producing sufficient quantities for model training still requires significant effort. 
Therefore, we augment these human-written scenarios through few-shot in-context learning~\citep{NEURIPS2020_1457c0d6}. 
As this augmentation process is optional, details are in 
Appendix~\ref{sec:scenario_augmentation}.



\paragraph{Dialogue Generation}
Formally, when a language agent $\mathcal{M}_a$ processes user input $x_T$ at each turn $T$, it leverages tools in a series of interactions $\mathcal{C}_T$, and finally produces a response $y_T$.
The agent also takes previous tool results and conversation history, it can be represented as
\begin{equation}
\label{eq:agent}
\left(\mathcal{C}_T, y_T\right) \sim \mathcal{M}_a\left(\mathcal{C}, y \mid x_T, \left[\left(x_t, \mathcal{C}_t, y_t\right)\right]_{t=1}^{T-1}\right).
\end{equation}

In order to distill such agent capabilities to small open-source LLMs, a multi-turn tool-using dialogue dataset $D_a=[(x_t^{(n)}, \mathcal{C}_t^{(n)}, y_t^{(n)})_{t=1}^T]_{n=1}^N$ is required.
To generate realistic user inputs $x_t$,
we employ persona-equipped LLMs~\citep{tseng-etal-2024-two} that simulate user behavior based on predefined scenarios.
The scenario data consist of $N$ user scenarios $D_u = \{s^{(n)},x_1^{(n)}\}_{n=1}^N$, where each includes user information $s$, and initial query $x_1$.
For the first turn, we generate a response $y_1$ from the target agent using Equation~\ref{eq:agent} and $x_1$.
For subsequent $T$-turns, we employ a user simulator $\mathcal{M}_u$ to generate new queries based on the scenario and previous conversation history in
\begin{equation}
\label{eq:user}
x_T \sim \mathcal{M}_u\left(x \mid s, \left[\left(x_t, y_t\right)\right]_{t=1}^{T-1}\right).
\end{equation}
The target agent is unaware that the user is a simulated by another LLM and cannot access the user scenario that guides the queries.

Following \citeposs{li2023camel} work, we sample multi-turn dialogue between the target agent $\mathcal{M}_a$ and user simulator $\mathcal{M}_u$.
Through this process, we effectively transform the user scenario dataset $D_u$ into multi-turn tool-using dialogue dataset $D_a$.
Once $D_a$ is generated, we fine-tune an open-source LLM by SFT manner to distill the target language agent capability leveraged by state-of-the-art LLMs.
It is noteworthy that this process is agnostic to the target agent architecture and can be generally applied to any agent structure.
In addition, for agents like Tox-chat that employ separate LLMs as tool, \ie summary and QA LLMs,
we record the inputs and outputs of these LLMs during the dialogue generation and use them as training data along with the dialogue data.
Appendix~\ref{sec:appendix-generated-data} provides examples of the generated scenarios and dialogues.

\section{Experiment}
For the experiments, we collect 100 human-written scenarios and divide them equally into training and evaluation sets.
The 50 training scenarios are then augmented using GPT-4o~\cite{openai-2024-gpt4o}, generating 971 multi-turn tool-using dialogues, 972 summarization pairs (for \texttt{BM25\_Search}), and 2484 QA pairs (for \texttt{QA\_Specific}).

We select two variants from the Llama 3 family~\citep{grattafiori2024llama3herdmodels}: Llama-3.1 8B (Llama-8B) and Llama-3.2 1B (Llama-1B).
Each model is fine-tuned on all three types of generated data (dialogues, summary pairs, and QA pairs) simultaneously, enabling a single fine-tuned model to perform all three required tasks.
For detailed information on data generation and fine-tuning settings, please refer to Appendix~\ref{sec:training_detail}.

\subsection{Baselines and Evaluation Metrics}
To evaluate the effectiveness of our Tox-chat architecture and the distillation process, we compare our models against two baselines: (1) vanilla GPT and Llama models without tool usage, and (2) GPT and Llama models with tool access but no fine-tuning.
Note that Tox-chat with GPT backbone serves as the upper bound of our fine-tuned model, as it generated the supervision data.
Evaluation is based on two LLM-as-a-judge metrics~\cite{zheng2023judging}, along with one retrieval-based metric.
For full judge system prompt and input format, please refer to Appendix~\ref{sec:prompts}.
Agreement between LLM-as-a-judge and human evaluations is reported in Appendix~\ref{sec:human_agreement}.

\paragraph{DB Consistency}
The judge model assesses whether the agent’s response is consistent with the Tox-info database. We provide the judge with up to six chemical documents relevant to the user's query (two on average), and measure percentage of `Yes' verdicts.

\paragraph{Preference}
Each model is compared pairwise against vanilla Llama-8B. The judge evaluates both models' responses and selects better response in terms of helpfulness, relevance, etc. Here, we use two-turn dialogue to capture conversational depth. To mitigate position bias, we swap the order of the responses in each scenario and average the results.

\paragraph{Search Success Rate}
The proportion of cases where the Tox-chat architecture successfully retrieves at least one reference document using BM25 or hierarchical search.
The reference documents are identical to those used in the DB Consistency evaluation.

\subsection{Experimental Results}
As shown in Table~\ref{tab:main-result}, adding the Tox-chat architecture significantly improves DB consistency across all backbones.

However, we observe a slight decline in preference scores when applying the Tox-chat architecture to commercial LLMs, which is partly attributed to the \emph{DB-only} constraint.  
Since Tox-chat is explicitly restricted from generating content outside the database, it tends to have limitations in fluency and coverage compared to other models that leverage internal knowledge.
Nevertheless, Tox-chat achieves high search success rate and DB Consistency, and 
the average response length is also relatively short, indicating that the system effectively retrieves and utilizes essential information without unnecessary elaboration.
It is also worth noting that baselines generating longer responses may have received higher preference scores due to the known bias of LLM-as-a-judge toward verbosity~\cite{saito2023verbosity}.

Despite these constraints, our fine-tuned Tox-chat-8B model outperforms vanilla Llama-8B in both DB consistency and preference.  
Notably, it even demonstrates higher DB consistency than GPT-4o, highlighting the effectiveness of domain-specific grounding over general-purpose commercial LLMs.  
Qualitative comparisons are provided in Appendix~\ref{sec:qual-results}.

\subsection{Ablation Study}
\label{sec:ablation}
We conduct an ablation study to demonstrate the effectiveness of both the Tox-chat architecture and our scenario-based dialogue generation approach. 
For these experiments, we maintain the same experimental settings as before, except that we use GPT-4o-mini as the backbone for cost-efficient dialogue generation.

Fig.~\ref{fig:token_count} illustrates the changes in token length across different Tox-chat architecture configurations.
"Full Doc." shows the average token count when using the entire document without a hierarchical structure from \texttt{Read General} to \texttt{QA Specific}. 
"RAG" displays the average token count when simply inserting all retrieved sections without additional searches nor summarization, similar to conventional RAG approaches.
In both cases, we observe a significant increase in tool output tokens (approximately 5.8x and 5.6x respectively compared to our proposed model), which dramatically increases GPU memory requirements during training.
The figure also shows the cases when our model does not use Summary LLM (w/o Summary) or QA LLM (w/o QA).
In both configurations, the average token count increases, demonstrating that each LLM component plays a definitive role in reducing context length.



\begin{figure}[t]
  \centering
  \begin{tikzpicture}
    \begin{axis}[
      xbar stacked,
      width=0.85\linewidth,
      symbolic y coords={Full Doc.,RAG,w/o Summary,w/o QA,\textbf{Ours}},
      xtick={0, 2000, 4000, 6000, 8000, 10000, 12000},
      ytick=data,
      scaled x ticks=false,
      legend style={at={(0.3,1.05)}, anchor=south, legend columns=3},
      xticklabel={\pgfmathparse{\tick/1000}\pgfmathprintnumber[fixed,precision=0]{\pgfmathresult}k},
      label style={font=\footnotesize},
      height=0.6\linewidth
    ]
      \addplot+[fill=blue!40] coordinates {
        (2401.0,Full Doc.)
        (1604.0,RAG)
        (2672.0,w/o Summary)
        (2636.0,w/o QA)
        (2686.0,\textbf{Ours})
      };
      \addplot+[fill=red!40] coordinates {
        (94.653964984552,Full Doc.)
        (92.25849639546858,RAG)
        (91.39031925849639,w/o Summary)
        (93.27497425334707,w/o QA)
        (91.67662203913491,\textbf{Ours})
      };
      \addplot+[fill=green!40] coordinates {
        (229.44593202883624,Full Doc.)
        (90.58187435633367,RAG)
        (357.0401647785788,w/o Summary)
        (347.42842430484035,w/o QA)
        (398.8259526261586,\textbf{Ours})
      };
      \addplot+[fill=orange!40] coordinates {
        (9289.811534500515,Full Doc.)
        (8885.4438722966,RAG)
        (6674.546858908342,w/o Summary)
        (2516.397528321318,w/o QA)
        (1589.223480947477,\textbf{Ours})
      };
      \addplot+[fill=purple!40] coordinates {
        (778.1050463439752,Full Doc.)
        (716.3542739443873,RAG)
        (579.3110195674562,w/o Summary)
        (727.0556127703398,w/o QA)
        (580.8053553038105,\textbf{Ours})
      };
  
      \legend{System, User, Tool Call, Tool Output, Response}
    \end{axis}
  \end{tikzpicture}   
  \caption{Average number of tokens per multi-turn dialogue generated by the teacher agent (distillation data) across different architecture configurations.}
  \label{fig:token_count}
\end{figure}


\begin{table}[t]
  \centering
  \resizebox{.9\columnwidth}{!}{%
  \begin{tabular}{ll|cc}
    \toprule
    \textbf{Abl.} & \textbf{Model}  & \textbf{Cons.}   & \textbf{\begin{tabular}[c]{@{}c@{}}Preference\\ (W/L/D)\end{tabular}} \\
    \midrule
    \textbf{Ours} &    & \textbf{70}       & 36 / 22 / 42 \\ 
    \midrule
    Arch. & w/o BM25          & 64       & 30 / 32 / 38 \\
          & w/o Sec. Search   & 54       & 36 / 26 / 38 \\
    \midrule
    Data  & Material        & 64       & 24 / 26 / 50 \\
          & Question        & 64       & 22 / 50 / 28 \\
          & Single-turn     & 60       & \textbf{38 / 30 / 32} \\
    \bottomrule
  \end{tabular}
  }
  \caption{Ablation studies on architecture and dialogue data generation. 
All models in this table are fine-tuned. 
For example, arch. w/o BM25 means that dialogues were generated without the BM25 tool, and Llama was then fine-tuned on those dialogues.}
  \label{tab:ablation}
  \vspace{-0.5em}
\end{table}

We evaluate our models using two key metrics: database consistency (Cons.) and human preference (Preference).
The upper part of Table~\ref{tab:ablation} presents these metrics for two architectural variations: when using only BM25 summarization without section search (w/o Sec. Search), similar to RECOMP~\citep{xu2024recomp}, and when using only section search without BM25 retrieval (w/o BM25). 
Both cases result in shorter average token counts than our model ("w/o Sec. Search" reduces tokens by approximately 33\% and "w/o BM25" by approximately 16\%). 
However, without section search, database consistency significantly decreases (54 vs. 70), and without BM25, the ability to provide comprehensive information is limited, leading to lower preference scores (30/32/38).

The lower part of the table shows metrics when collecting distillation data through methods other than scenario-based multi-turn dialogue generation.
"Material" represents cases where the user simulator continues conversations with only a list of substance names without scenarios, similar to scenario augmentation.
"Question" shows cases where the user simulator continues dialogues using only the given initial query without utilizing the user's intent or situation from scenario data.
Both cases fail to effectively simulate users during multi-turn generation, resulting in deteriorated preference scores, with "Question" showing a particularly high loss rate (50\%).
Meanwhile, "Single-turn" shows the results of a model trained only with single-turn conversations.
While this approach maintains relatively good preference scores (38/30/32) due to training on human-written queries exclusively, it shows a notable decrease in database consistency (60 vs. 70), likely due to the lack of learning various interactions in multi-turn processes.
These results demonstrate the effectiveness of scenario-based data when generating multi-turn dialogues that simulate real users.

\subsection{User Study}
To evaluate Tox-chat comprehensively, we conducted a user study with 14 participants.
Seven had academic or professional backgrounds in the chemical or food industries, while the other seven possessed general domain knowledge in toxicology.  
Prior to the main experiment, all participants were instructed to search and read specific documents on the Tox-info website to familiarize themselves with the database.

Participants interacted with three open-source-based chatbots: Tox-chat-8B, Llama-8B with tool access, and vanilla Llama-8B.  
They were asked to evaluate which model performed best according to two criteria:  
(1) how well the model understood the user’s query and responded naturally, and  
(2) the perceived accuracy and reliability of its responses.

As shown in Fig.~\ref{fig:user-study}, Tox-chat achieved the highest user preference scores across both criteria among the open-source models.  
Participants noted that Tox-chat-8B was more effective in leveraging the Tox-info database, delivering concise and relevant answers.  
They also highlighted that its clear grounding in source documents contributed to higher perceived reliability.

\begin{figure}[t]
    \centering
\begin{tikzpicture}
\begin{axis}[
    ybar,
    bar width=18pt,
    width=1.05\linewidth,
    height=4.4cm, 
    ymin=0, ymax=11,
    enlarge x limits=0.25,
    ylabel={User Preference},
    ylabel style={font=\footnotesize, at={(axis description cs:-0.08,0.5)}, anchor=south},
    xlabel={},
    symbolic x coords={Tox-chat-8B, Tox-chat [Llama-8B], Llama-8B},
    xtick=data,
    xticklabels={
      {Tox-chat-8B},
      {Tox-chat\\\textnormal{[Llama-8B]}}, 
      {Llama-8B}
    },
    tick label style={font=\footnotesize, align=center},
    label style={font=\footnotesize},
    nodes near coords,
    nodes near coords style={font=\footnotesize, yshift=2pt},
    legend style={
        font=\footnotesize,
        at={(0.98,0.98)},
        anchor=north east,
        draw=black,
        fill=white,
        inner sep=4pt
    },
    legend cell align={left},
    axis lines*=left,
    clip=true, 
    axis on top  
]

\addplot+[ybar,fill=blue!50,draw=black] 
    coordinates {(Tox-chat-8B,9) (Tox-chat [Llama-8B],3) (Llama-8B,2)};
\addlegendentry{Response Naturalness}

\addplot+[ybar,fill=red!50,draw=black] 
    coordinates {(Tox-chat-8B,10) (Tox-chat [Llama-8B],4) (Llama-8B,0)};
\addlegendentry{Accuracy \& Reliability}

\end{axis}
\end{tikzpicture}
    \caption{User study results showing participant preferences based on response naturalness and accuracy/reliability criteria.}
    \label{fig:user-study}
    \vspace{-0.5em}
\end{figure}

To further compare user experience with GPT-4o, the same participants were asked to use both Tox-chat-8B and GPT-4o side-by-side in a setting similar to the previous experiment.
Compared to GPT-4o, participants found our Tox-chat-8B demonstrated superior reliability and practical utility for professional use.
For example, when queried about methanol toxicity, GPT-4o provided only general explanations, whereas Tox-chat-8B delivered comprehensive details including LD50 values, specific sources, and experimental conditions. 
When tested with ``gallium hydroxide'', a non-existent compound, GPT-4o generated hallucinations as if the compound existed, while Tox-chat-8B correctly identified the error and redirected to ``gallium trichloride'', an actual compound. 
Based on these observations, participants concluded that Tox-chat-8B represents a more reliable agent for expert applications.
Detailed descriptions of the user study setup and analysis are provided in Appendix~\ref{sec:appendix-user-study}.

\section{Conclusion}

Our work successfully demonstrates an instance of effective language agents in resource-constrained environments,
specifically addressing the challenges of limited Korean toxicity information data and computational resources. 
By sharing these experiences, we anticipate to provide valuable guidance to researchers and developers facing similar resource constraints in specialized domains.

\section*{Ethical Considerations}
\label{sec:ethical_consideration}
Safety and hallucination are critical concerns in toxicology information services, where factual accuracy is essential and speculative content can pose real-world risks.
While proprietary LLMs may reason safely over retrieved documents, our system targets smaller distilled models where fine-tuning or distillation often compromises safety~\cite{qi2024finetuning} and reasoning capabilities~\cite{pmlr-v202-fu23d}, increasing the risk of hallucination.

Before the main experiments, we have trained an LLM on dialogues generated without any constraint.
In user evaluations, this model often relied on internal knowledge and produced free-form answers, making it unsuitable for deployment.
Based on this qualitative feedback, we impose a strict \emph{DB-only} constraint: the model must generate responses grounded solely in the content retrieved from Tox-info.
The system prompt explicitly (1) prohibits generation beyond retrieved evidence, and (2) discourages reliance on internal knowledge or reasoning.
This constraint not only enhances safety but also reduces hallucination during distillation by reducing the reasoning burden.

Nevertheless, the current model has not undergone safety-specific fine-tuning.
The scenario data used in this research does not cover malicious or adversarial user interactions;
instead, it is excluscively constructed around legitimate use cases.
One potential direction for future work is to introduce safety-aware training through the generation of adversarial or misuse-oriented dialogue data, in the spirit of red teaming approaches~\citep{perez-etal-2022-red}.
We leave such safety-focused augmentation and evaluation to future research.

\section*{Limitations}
\label{sec:limitations}
The \emph{DB-only} constraint clearly enhances safety and reduces hallucinations but at the cost of reduced fluency and coverage, occasionally lowering preference scores.
We adopt this restriction as a practical requirement for toxicology applications, yet it entails an inevitable trade-off.
In particular, this constraint prevents the model from performing reasoning beyond retrieval, so Tox-chat functions primarily as a retriever and summarizer rather than a fully capable language agent.
To develop more advanced agents, future research will need methods that can safely incorporate internal reasoning abilities while preserving reliability.

\section*{Acknowledgments}
This work was supported by Institute for Information \& communications Technology Planning \& Evaluation(IITP) grant funded by the Korea government(MSIT) (RS-2019-II190075, Artificial Intelligence Graduate School Program(KAIST)), the National Research Foundation of Korea(NRF) grant funded by the Korea government(MSIT) (No. RS-2025-00555621), and Artificial intelligence industrial convergence cluster development project funded by the Ministry of Science and ICT(MSIT, Korea) \& Gwangju Metropolitan City.

\bibliography{anthology,custom}

\begin{thebibliography}{43}
\providecommand{\natexlab}[1]{#1}

\bibitem[{Achiam et~al.(2024)Achiam, Adler, Agarwal, Ahmad, Akkaya, Aleman,
  Almeida, Altenschmidt, Altman, Anadkat, and et~al.}]{openai-2024-gpt}
Josh Achiam, Steven Adler, Sandhini Agarwal, Lama Ahmad, Ilge Akkaya,
  Florencia~Leoni Aleman, Diogo Almeida, Janko Altenschmidt, Sam Altman,
  Shyamal Anadkat, and et~al. 2024.
\newblock \href {https://arxiv.org/abs/2303.08774} {Gpt-4 technical report}.
\newblock \emph{Preprint}, arXiv:2303.08774.

\bibitem[{Asai et~al.(2024)Asai, Wu, Wang, Sil, and
  Hajishirzi}]{asai2024selfrag}
Akari Asai, Zeqiu Wu, Yizhong Wang, Avirup Sil, and Hannaneh Hajishirzi. 2024.
\newblock \href {https://openreview.net/forum?id=hSyW5go0v8} {Self-{RAG}:
  Learning to retrieve, generate, and critique through self-reflection}.
\newblock In \emph{The Twelfth International Conference on Learning
  Representations}.

\bibitem[{Brown et~al.(2020)Brown, Mann, Ryder, Subbiah, Kaplan, Dhariwal,
  Neelakantan, Shyam, Sastry, Askell, Agarwal, Herbert-Voss, Krueger, Henighan,
  Child, Ramesh, Ziegler, Wu, Winter, Hesse, Chen, Sigler, Litwin, Gray, Chess,
  Clark, Berner, McCandlish, Radford, Sutskever, and
  Amodei}]{NEURIPS2020_1457c0d6}
Tom Brown, Benjamin Mann, Nick Ryder, Melanie Subbiah, Jared~D Kaplan, Prafulla
  Dhariwal, Arvind Neelakantan, Pranav Shyam, Girish Sastry, Amanda Askell,
  Sandhini Agarwal, Ariel Herbert-Voss, Gretchen Krueger, Tom Henighan, Rewon
  Child, Aditya Ramesh, Daniel Ziegler, Jeffrey Wu, Clemens Winter, Chris
  Hesse, Mark Chen, Eric Sigler, Mateusz Litwin, Scott Gray, Benjamin Chess,
  Jack Clark, Christopher Berner, Sam McCandlish, Alec Radford, Ilya Sutskever,
  and Dario Amodei. 2020.
\newblock \href
  {https://proceedings.neurips.cc/paper_files/paper/2020/file/1457c0d6bfcb4967418bfb8ac142f64a-Paper.pdf}
  {Language models are few-shot learners}.
\newblock In \emph{Advances in Neural Information Processing Systems},
  volume~33, pages 1877--1901. Curran Associates, Inc.

\bibitem[{Chen et~al.(2023)Chen, Shu, Shareghi, Collier, Narasimhan, and
  Yao}]{chen2023fireactlanguageagentfinetuning}
Baian Chen, Chang Shu, Ehsan Shareghi, Nigel Collier, Karthik Narasimhan, and
  Shunyu Yao. 2023.
\newblock \href {https://arxiv.org/abs/2310.05915} {Fireact: Toward language
  agent fine-tuning}.
\newblock \emph{Preprint}, arXiv:2310.05915.

\bibitem[{Chiang et~al.(2023)Chiang, Li, Lin, Sheng, Wu, Zhang, Zheng, Zhuang,
  Zhuang, Gonzalez, Stoica, and Xing}]{vicuna2023}
Wei-Lin Chiang, Zhuohan Li, Zi~Lin, Ying Sheng, Zhanghao Wu, Hao Zhang, Lianmin
  Zheng, Siyuan Zhuang, Yonghao Zhuang, Joseph~E. Gonzalez, Ion Stoica, and
  Eric~P. Xing. 2023.
\newblock \href {https://lmsys.org/blog/2023-03-30-vicuna/} {Vicuna: An
  open-source chatbot impressing gpt-4 with 90\%* chatgpt quality}.

\bibitem[{Fu et~al.(2023)Fu, Peng, Ou, Sabharwal, and Khot}]{pmlr-v202-fu23d}
Yao Fu, Hao Peng, Litu Ou, Ashish Sabharwal, and Tushar Khot. 2023.
\newblock \href {https://proceedings.mlr.press/v202/fu23d.html} {Specializing
  smaller language models towards multi-step reasoning}.
\newblock In \emph{Proceedings of the 40th International Conference on Machine
  Learning}, volume 202 of \emph{Proceedings of Machine Learning Research},
  pages 10421--10430. PMLR.

\bibitem[{Gao et~al.(2024)Gao, Xiong, Gao, Jia, Pan, Bi, Dai, Sun, Wang, and
  Wang}]{gao-etal-2024-retrieval}
Yunfan Gao, Yun Xiong, Xinyu Gao, Kangxiang Jia, Jinliu Pan, Yuxi Bi, Yi~Dai,
  Jiawei Sun, Meng Wang, and Haofen Wang. 2024.
\newblock \href {https://arxiv.org/abs/2312.10997} {Retrieval-augmented
  generation for large language models: A survey}.
\newblock \emph{Preprint}, arXiv:2312.10997.

\bibitem[{Grattafiori et~al.(2024)Grattafiori, Dubey, Jauhri, Pandey, Kadian,
  Al-Dahle, Letman, Mathur, Schelten, Vaughan, and
  et~al.}]{grattafiori2024llama3herdmodels}
Aaron Grattafiori, Abhimanyu Dubey, Abhinav Jauhri, Abhinav Pandey, Abhishek
  Kadian, Ahmad Al-Dahle, Aiesha Letman, Akhil Mathur, Alan Schelten, Alex
  Vaughan, and et~al. 2024.
\newblock \href {https://arxiv.org/abs/2407.21783} {The llama 3 herd of
  models}.
\newblock \emph{Preprint}, arXiv:2407.21783.

\bibitem[{Gugger et~al.(2022)Gugger, Debut, Wolf, Schmid, Mueller, Mangrulkar,
  Sun, and Bossan}]{accelerate}
Sylvain Gugger, Lysandre Debut, Thomas Wolf, Philipp Schmid, Zachary Mueller,
  Sourab Mangrulkar, Marc Sun, and Benjamin Bossan. 2022.
\newblock Accelerate: Training and inference at scale made simple, efficient
  and adaptable.
\newblock \url{https://github.com/huggingface/accelerate}.

\bibitem[{Hsieh et~al.(2023)Hsieh, Li, Yeh, Nakhost, Fujii, Ratner, Krishna,
  Lee, and Pfister}]{hsieh-etal-2023-distilling}
Cheng-Yu Hsieh, Chun-Liang Li, Chih-kuan Yeh, Hootan Nakhost, Yasuhisa Fujii,
  Alex Ratner, Ranjay Krishna, Chen-Yu Lee, and Tomas Pfister. 2023.
\newblock \href {https://doi.org/10.18653/v1/2023.findings-acl.507} {Distilling
  step-by-step! outperforming larger language models with less training data
  and smaller model sizes}.
\newblock In \emph{Findings of the Association for Computational Linguistics:
  ACL 2023}, pages 8003--8017, Toronto, Canada. Association for Computational
  Linguistics.

\bibitem[{Hurst et~al.(2024)Hurst, Lerer, Goucher, Perelman, Ramesh, Clark,
  Ostrow, Welihinda, Hayes, Radford, and et~al.}]{openai-2024-gpt4o}
Aaron Hurst, Adam Lerer, Adam~P. Goucher, Adam Perelman, Aditya Ramesh, Aidan
  Clark, AJ~Ostrow, Akila Welihinda, Alan Hayes, Alec Radford, and et~al. 2024.
\newblock \href {https://arxiv.org/abs/2410.21276} {Gpt-4o system card}.
\newblock \emph{Preprint}, arXiv:2410.21276.

\bibitem[{Jeong et~al.(2024)Jeong, Baek, Cho, Hwang, and
  Park}]{jeong-etal-2024-adaptive}
Soyeong Jeong, Jinheon Baek, Sukmin Cho, Sung~Ju Hwang, and Jong Park. 2024.
\newblock \href {https://doi.org/10.18653/v1/2024.naacl-long.389}
  {Adaptive-{RAG}: Learning to adapt retrieval-augmented large language models
  through question complexity}.
\newblock In \emph{Proceedings of the 2024 Conference of the North American
  Chapter of the Association for Computational Linguistics: Human Language
  Technologies (Volume 1: Long Papers)}, pages 7036--7050, Mexico City, Mexico.
  Association for Computational Linguistics.

\bibitem[{Kendall(1938)}]{kendall1938new}
Maurice~G Kendall. 1938.
\newblock A new measure of rank correlation.
\newblock \emph{Biometrika}, 30(1/2):81--93.

\bibitem[{Li et~al.(2023{\natexlab{a}})Li, Hammoud, Itani, Khizbullin, and
  Ghanem}]{li2023camel}
Guohao Li, Hasan Abed Al~Kader Hammoud, Hani Itani, Dmitrii Khizbullin, and
  Bernard Ghanem. 2023{\natexlab{a}}.
\newblock \href {https://openreview.net/forum?id=3IyL2XWDkG} {{CAMEL}:
  Communicative agents for ''mind'' exploration of large language model
  society}.
\newblock In \emph{Thirty-seventh Conference on Neural Information Processing
  Systems}.

\bibitem[{Li et~al.(2023{\natexlab{b}})Li, Zhao, Yu, Song, Li, Yu, Li, Huang,
  and Li}]{li-etal-2023-api}
Minghao Li, Yingxiu Zhao, Bowen Yu, Feifan Song, Hangyu Li, Haiyang Yu, Zhoujun
  Li, Fei Huang, and Yongbin Li. 2023{\natexlab{b}}.
\newblock \href {https://doi.org/10.18653/v1/2023.emnlp-main.187} {{API}-bank:
  A comprehensive benchmark for tool-augmented {LLM}s}.
\newblock In \emph{Proceedings of the 2023 Conference on Empirical Methods in
  Natural Language Processing}, pages 3102--3116, Singapore. Association for
  Computational Linguistics.

\bibitem[{Li et~al.(2024)Li, Zhao, Chia, Ding, Joty, Poria, and
  Bing}]{li2024chainofknowledge}
Xingxuan Li, Ruochen Zhao, Yew~Ken Chia, Bosheng Ding, Shafiq Joty, Soujanya
  Poria, and Lidong Bing. 2024.
\newblock \href {https://openreview.net/forum?id=cPgh4gWZlz}
  {Chain-of-knowledge: Grounding large language models via dynamic knowledge
  adapting over heterogeneous sources}.
\newblock In \emph{The Twelfth International Conference on Learning
  Representations}.

\bibitem[{Liu et~al.(2025)Liu, Feng, Xue, Wang, Wu, Lu, Zhao, Deng, Zhang,
  Ruan, and et~al.}]{deepseekai-2025-deepseek}
Aixin Liu, Bei Feng, Bing Xue, Bingxuan Wang, Bochao Wu, Chengda Lu, Chenggang
  Zhao, Chengqi Deng, Chenyu Zhang, Chong Ruan, and et~al. 2025.
\newblock \href {https://arxiv.org/abs/2412.19437} {Deepseek-v3 technical
  report}.
\newblock \emph{Preprint}, arXiv:2412.19437.

\bibitem[{Lo et~al.(2023)Lo, Sridhar, Xu, Zhu, and
  Zhou}]{lo-etal-2023-hierarchical}
Robert Lo, Abishek Sridhar, Frank Xu, Hao Zhu, and Shuyan Zhou. 2023.
\newblock \href {https://doi.org/10.18653/v1/2023.findings-emnlp.685}
  {Hierarchical prompting assists large language model on web navigation}.
\newblock In \emph{Findings of the Association for Computational Linguistics:
  EMNLP 2023}, pages 10217--10244, Singapore. Association for Computational
  Linguistics.

\bibitem[{Loshchilov and Hutter(2019)}]{loshchilov2018decoupled}
Ilya Loshchilov and Frank Hutter. 2019.
\newblock \href {https://openreview.net/forum?id=Bkg6RiCqY7} {Decoupled weight
  decay regularization}.
\newblock In \emph{International Conference on Learning Representations}.

\bibitem[{Luo et~al.(2024)Luo, Liu, Xiao, Zhou, Chen, Zhao, and
  Liu}]{luo-etal-2024-landmark}
Kun Luo, Zheng Liu, Shitao Xiao, Tong Zhou, Yubo Chen, Jun Zhao, and Kang Liu.
  2024.
\newblock \href {https://doi.org/10.18653/v1/2024.acl-long.180} {Landmark
  embedding: A chunking-free embedding method for retrieval augmented
  long-context large language models}.
\newblock In \emph{Proceedings of the 62nd Annual Meeting of the Association
  for Computational Linguistics (Volume 1: Long Papers)}, pages 3268--3281,
  Bangkok, Thailand. Association for Computational Linguistics.

\bibitem[{Peng et~al.(2023)Peng, Li, He, Galley, and Gao}]{peng2023instruction}
Baolin Peng, Chunyuan Li, Pengcheng He, Michel Galley, and Jianfeng Gao. 2023.
\newblock Instruction tuning with gpt-4.
\newblock \emph{arXiv preprint arXiv:2304.03277}.

\bibitem[{Perez et~al.(2022)Perez, Huang, Song, Cai, Ring, Aslanides, Glaese,
  McAleese, and Irving}]{perez-etal-2022-red}
Ethan Perez, Saffron Huang, Francis Song, Trevor Cai, Roman Ring, John
  Aslanides, Amelia Glaese, Nat McAleese, and Geoffrey Irving. 2022.
\newblock \href {https://doi.org/10.18653/v1/2022.emnlp-main.225} {Red teaming
  language models with language models}.
\newblock In \emph{Proceedings of the 2022 Conference on Empirical Methods in
  Natural Language Processing}, pages 3419--3448, Abu Dhabi, United Arab
  Emirates. Association for Computational Linguistics.

\bibitem[{Qi et~al.(2024)Qi, Zeng, Xie, Chen, Jia, Mittal, and
  Henderson}]{qi2024finetuning}
Xiangyu Qi, Yi~Zeng, Tinghao Xie, Pin-Yu Chen, Ruoxi Jia, Prateek Mittal, and
  Peter Henderson. 2024.
\newblock \href {https://openreview.net/forum?id=hTEGyKf0dZ} {Fine-tuning
  aligned language models compromises safety, even when users do not intend
  to!}
\newblock In \emph{The Twelfth International Conference on Learning
  Representations}.

\bibitem[{Qin et~al.(2024)Qin, Liang, Ye, Zhu, Yan, Lu, Lin, Cong, Tang, Qian,
  Zhao, Hong, Tian, Xie, Zhou, Gerstein, dahai li, Liu, and
  Sun}]{qin2024toolllm}
Yujia Qin, Shihao Liang, Yining Ye, Kunlun Zhu, Lan Yan, Yaxi Lu, Yankai Lin,
  Xin Cong, Xiangru Tang, Bill Qian, Sihan Zhao, Lauren Hong, Runchu Tian,
  Ruobing Xie, Jie Zhou, Mark Gerstein, dahai li, Zhiyuan Liu, and Maosong Sun.
  2024.
\newblock \href {https://openreview.net/forum?id=dHng2O0Jjr} {Tool{LLM}:
  Facilitating large language models to master 16000+ real-world {API}s}.
\newblock In \emph{The Twelfth International Conference on Learning
  Representations}.

\bibitem[{Robertson and Zaragoza(2009)}]{robertson-etal-2009-probabilistic}
Stephen Robertson and Hugo Zaragoza. 2009.
\newblock \href {https://doi.org/10.1561/1500000019} {The probabilistic
  relevance framework: Bm25 and beyond}.
\newblock \emph{Found. Trends Inf. Retr.}, 3(4):333–389.

\bibitem[{Saito et~al.(2023)Saito, Wachi, Wataoka, and
  Akimoto}]{saito2023verbosity}
Keita Saito, Akifumi Wachi, Koki Wataoka, and Youhei Akimoto. 2023.
\newblock Verbosity bias in preference labeling by large language models.
\newblock \emph{arXiv preprint arXiv:2310.10076}.

\bibitem[{Shim et~al.(2025)Shim, Seo, Lim, and Jo}]{shim2025tooldial}
Jeonghoon Shim, Gyuhyeon Seo, Cheongsu Lim, and Yohan Jo. 2025.
\newblock \href {https://openreview.net/forum?id=J1J5eGJsKZ} {Tooldial:
  Multi-turn dialogue generation method for tool-augmented language models}.
\newblock In \emph{The Thirteenth International Conference on Learning
  Representations}.

\bibitem[{Shumailov et~al.(2024)Shumailov, Shumaylov, Zhao, Papernot, Anderson,
  and Gal}]{Shumailov2024}
Ilia Shumailov, Zakhar Shumaylov, Yiren Zhao, Nicolas Papernot, Ross Anderson,
  and Yarin Gal. 2024.
\newblock \href {https://doi.org/10.1038/s41586-024-07566-y} {Ai models
  collapse when trained on recursively generated data}.
\newblock \emph{Nature}, 631(8022):755--759.

\bibitem[{Su et~al.(2024)Su, Yang, Yao, and Yu}]{su-etal-2024-language}
Yu~Su, Diyi Yang, Shunyu Yao, and Tao Yu. 2024.
\newblock \href {https://doi.org/10.18653/v1/2024.emnlp-tutorials.3} {Language
  agents: Foundations, prospects, and risks}.
\newblock In \emph{Proceedings of the 2024 Conference on Empirical Methods in
  Natural Language Processing: Tutorial Abstracts}, pages 17--24, Miami,
  Florida, USA. Association for Computational Linguistics.

\bibitem[{Tan and Jiang(2023)}]{DBLP:journals/debu/Tan023}
Zhaoxuan Tan and Meng Jiang. 2023.
\newblock \href {http://sites.computer.org/debull/A23dec/p57.pdf} {User
  modeling in the era of large language models: Current research and future
  directions}.
\newblock \emph{{IEEE} Data Eng. Bull.}, 46(4):57--96.

\bibitem[{Taori et~al.(2023)Taori, Gulrajani, Zhang, Dubois, Li, Guestrin,
  Liang, and Hashimoto}]{alpaca}
Rohan Taori, Ishaan Gulrajani, Tianyi Zhang, Yann Dubois, Xuechen Li, Carlos
  Guestrin, Percy Liang, and Tatsunori~B. Hashimoto. 2023.
\newblock Stanford alpaca: An instruction-following llama model.
\newblock \url{https://github.com/tatsu-lab/stanford_alpaca}.

\bibitem[{Tseng et~al.(2024)Tseng, Huang, Hsiao, Chen, Huang, Meng, and
  Chen}]{tseng-etal-2024-two}
Yu-Min Tseng, Yu-Chao Huang, Teng-Yun Hsiao, Wei-Lin Chen, Chao-Wei Huang,
  Yu~Meng, and Yun-Nung Chen. 2024.
\newblock \href {https://doi.org/10.18653/v1/2024.findings-emnlp.969} {Two
  tales of persona in {LLM}s: A survey of role-playing and personalization}.
\newblock In \emph{Findings of the Association for Computational Linguistics:
  EMNLP 2024}, pages 16612--16631, Miami, Florida, USA. Association for
  Computational Linguistics.

\bibitem[{Wang et~al.(2024)Wang, Ma, Feng, Zhang, Yang, Zhang, Chen, Tang,
  Chen, Lin, Zhao, Wei, and Wen}]{wang-etal-2024-survey-large}
Lei Wang, Chen Ma, Xueyang Feng, Zeyu Zhang, Hao Yang, Jingsen Zhang, Zhiyuan
  Chen, Jiakai Tang, Xu~Chen, Yankai Lin, Wayne~Xin Zhao, Zhewei Wei, and
  Jirong Wen. 2024.
\newblock \href {https://doi.org/10.1007/s11704-024-40231-1} {A survey on large
  language model based autonomous agents}.
\newblock \emph{Frontiers of Computer Science}, 18(6).

\bibitem[{Wolf et~al.(2020)Wolf, Debut, Sanh, Chaumond, Delangue, Moi, Cistac,
  Rault, Louf, Funtowicz, Davison, Shleifer, von Platen, Ma, Jernite, Plu, Xu,
  Le~Scao, Gugger, Drame, Lhoest, and Rush}]{wolf-etal-2020-transformers}
Thomas Wolf, Lysandre Debut, Victor Sanh, Julien Chaumond, Clement Delangue,
  Anthony Moi, Pierric Cistac, Tim Rault, Remi Louf, Morgan Funtowicz, Joe
  Davison, Sam Shleifer, Patrick von Platen, Clara Ma, Yacine Jernite, Julien
  Plu, Canwen Xu, Teven Le~Scao, Sylvain Gugger, Mariama Drame, Quentin Lhoest,
  and Alexander Rush. 2020.
\newblock \href {https://doi.org/10.18653/v1/2020.emnlp-demos.6} {Transformers:
  State-of-the-art natural language processing}.
\newblock In \emph{Proceedings of the 2020 Conference on Empirical Methods in
  Natural Language Processing: System Demonstrations}, pages 38--45, Online.
  Association for Computational Linguistics.

\bibitem[{Xi et~al.(2023)Xi, Chen, Guo, He, Ding, Hong, Zhang, Wang, Jin, Zhou,
  Zheng, Fan, Wang, Xiong, Zhou, Wang, Jiang, Zou, Liu, Yin, Dou, Weng, Cheng,
  Zhang, Qin, Zheng, Qiu, Huang, and Gui}]{xi-etal-2023-rise}
Zhiheng Xi, Wenxiang Chen, Xin Guo, Wei He, Yiwen Ding, Boyang Hong, Ming
  Zhang, Junzhe Wang, Senjie Jin, Enyu Zhou, Rui Zheng, Xiaoran Fan, Xiao Wang,
  Limao Xiong, Yuhao Zhou, Weiran Wang, Changhao Jiang, Yicheng Zou, Xiangyang
  Liu, Zhangyue Yin, Shihan Dou, Rongxiang Weng, Wensen Cheng, Qi~Zhang,
  Wenjuan Qin, Yongyan Zheng, Xipeng Qiu, Xuanjing Huang, and Tao Gui. 2023.
\newblock \href {https://arxiv.org/abs/2309.07864} {The rise and potential of
  large language model based agents: A survey}.
\newblock \emph{Preprint}, arXiv:2309.07864.

\bibitem[{Xu et~al.(2024{\natexlab{a}})Xu, Shi, and Choi}]{xu2024recomp}
Fangyuan Xu, Weijia Shi, and Eunsol Choi. 2024{\natexlab{a}}.
\newblock \href {https://openreview.net/forum?id=mlJLVigNHp} {{RECOMP}:
  Improving retrieval-augmented {LM}s with context compression and selective
  augmentation}.
\newblock In \emph{The Twelfth International Conference on Learning
  Representations}.

\bibitem[{Xu et~al.(2024{\natexlab{b}})Xu, Ping, Wu, McAfee, Zhu, Liu,
  Subramanian, Bakhturina, Shoeybi, and Catanzaro}]{xu2024retrieval}
Peng Xu, Wei Ping, Xianchao Wu, Lawrence McAfee, Chen Zhu, Zihan Liu, Sandeep
  Subramanian, Evelina Bakhturina, Mohammad Shoeybi, and Bryan Catanzaro.
  2024{\natexlab{b}}.
\newblock \href {https://openreview.nt/forum?id=xw5nxFWMlo} {Retrieval meets
  long context large language models}.
\newblock In \emph{The Twelfth International Conference on Learning
  Representations}.

\bibitem[{Yin et~al.(2024)Yin, Brahman, Ravichander, Chandu, Chang, Choi, and
  Lin}]{yin-etal-2024-agent}
Da~Yin, Faeze Brahman, Abhilasha Ravichander, Khyathi Chandu, Kai-Wei Chang,
  Yejin Choi, and Bill~Yuchen Lin. 2024.
\newblock \href {https://doi.org/10.18653/v1/2024.acl-long.670} {Agent lumos:
  Unified and modular training for open-source language agents}.
\newblock In \emph{Proceedings of the 62nd Annual Meeting of the Association
  for Computational Linguistics (Volume 1: Long Papers)}, pages 12380--12403,
  Bangkok, Thailand. Association for Computational Linguistics.

\bibitem[{Zeng et~al.(2024)Zeng, Liu, Lu, Wang, Liu, Dong, and
  Tang}]{zeng-etal-2024-agenttuning}
Aohan Zeng, Mingdao Liu, Rui Lu, Bowen Wang, Xiao Liu, Yuxiao Dong, and Jie
  Tang. 2024.
\newblock \href {https://doi.org/10.18653/v1/2024.findings-acl.181}
  {{A}gent{T}uning: Enabling generalized agent abilities for {LLM}s}.
\newblock In \emph{Findings of the Association for Computational Linguistics:
  ACL 2024}, pages 3053--3077, Bangkok, Thailand. Association for Computational
  Linguistics.

\bibitem[{Zhang et~al.(2024)Zhang, Sun, Chen, Pfister, Zhang, and
  Arik}]{zhang2024chain}
Yusen Zhang, Ruoxi Sun, Yanfei Chen, Tomas Pfister, Rui Zhang, and Sercan~O
  Arik. 2024.
\newblock \href {https://openreview.net/forum?id=LuCLf4BJsr} {Chain of agents:
  Large language models collaborating on long-context tasks}.
\newblock In \emph{The Thirty-eighth Annual Conference on Neural Information
  Processing Systems}.

\bibitem[{Zhao et~al.(2024)Zhao, Zu, Hao, Lu, He, Ding, Gui, Zhang, and
  Huang}]{zhao-etal-2024-longagent}
Jun Zhao, Can Zu, Xu~Hao, Yi~Lu, Wei He, Yiwen Ding, Tao Gui, Qi~Zhang, and
  Xuanjing Huang. 2024.
\newblock \href {https://doi.org/10.18653/v1/2024.emnlp-main.912} {{LONGAGENT}:
  Achieving question answering for 128k-token-long documents through
  multi-agent collaboration}.
\newblock In \emph{Proceedings of the 2024 Conference on Empirical Methods in
  Natural Language Processing}, pages 16310--16324, Miami, Florida, USA.
  Association for Computational Linguistics.

\bibitem[{Zhao et~al.(2023)Zhao, Gu, Varma, Luo, Huang, Xu, Wright,
  Shojanazeri, Ott, Shleifer, Desmaison, Balioglu, Damania, Nguyen, Chauhan,
  Hao, Mathews, and Li}]{zhao2023pytorchfsdpexperiencesscaling}
Yanli Zhao, Andrew Gu, Rohan Varma, Liang Luo, Chien-Chin Huang, Min Xu, Less
  Wright, Hamid Shojanazeri, Myle Ott, Sam Shleifer, Alban Desmaison, Can
  Balioglu, Pritam Damania, Bernard Nguyen, Geeta Chauhan, Yuchen Hao, Ajit
  Mathews, and Shen Li. 2023.
\newblock \href {https://arxiv.org/abs/2304.11277} {Pytorch fsdp: Experiences
  on scaling fully sharded data parallel}.
\newblock \emph{Preprint}, arXiv:2304.11277.

\bibitem[{Zheng et~al.(2023)Zheng, Chiang, Sheng, Zhuang, Wu, Zhuang, Lin, Li,
  Li, Xing, Zhang, Gonzalez, and Stoica}]{zheng2023judging}
Lianmin Zheng, Wei-Lin Chiang, Ying Sheng, Siyuan Zhuang, Zhanghao Wu, Yonghao
  Zhuang, Zi~Lin, Zhuohan Li, Dacheng Li, Eric Xing, Hao Zhang, Joseph~E.
  Gonzalez, and Ion Stoica. 2023.
\newblock \href {https://openreview.net/forum?id=uccHPGDlao} {Judging
  {LLM}-as-a-judge with {MT}-bench and chatbot arena}.
\newblock In \emph{Thirty-seventh Conference on Neural Information Processing
  Systems Datasets and Benchmarks Track}.

\end{thebibliography}
\newpage
\appendix

\part*{Appendix}

\label{sec:appendix}
\section{Scenario Augmentation}
\label{sec:scenario_augmentation}
Since scenario diversity directly affects dialogue diversity, we employ practical methodologies to generate a wide range of scenarios.
Specifically, we sample 3-30 example scenarios from our human-written scenario pool and use them as examples for the LLM.
To enhance scenario diversity, we provide an additional list of chemical substance names that could be used in the scenario generation.
These names are derived from document titles within the Tox-info database.
To eliminate redundant scenarios, we filter out generated scenarios that have more than 15\% N-gram overlap with existing scenarios or previously generated ones.
The complete prompt used for scenario generation is provided in Appendix~\ref{sec:prompts}.

We also experimented with adding generated scenarios to the example set for subsequent generation.
However, using only human-written scenarios as examples results in higher diversity and fewer scenarios being removed by our N-gram filtering.
When LLM-generated scenarios constitute the majority of the example set, the LLM is more likely to generate similar outputs.
This phenomenon is similar to the output distribution collapse observed when models are repeatedly trained on LLM-generated text~\cite{Shumailov2024}

In practice, we utilize \texttt{gpt-4o-2024-11-20} as the backbone LLM.
We construct example scenarios based on 50 human-written scenarios and sample multiple times with a temperature of 0.5.
We repeatedly generate 1,000 scenarios at a time until the API usage cost exceeds \$20.
As a result, we generate 18,000 scenarios, of which 17,079 are filtered out using N-Gram similarity, resulting in 921 diverse scenarios.


\section{Tox-chat Training Detail}
\label{sec:training_detail}
\paragraph{Dialogue Generation}
Both the user simulator and the Tox-chat backbone for dialogue generation utilize \texttt{gpt-4o-2024-11-20}.
We combine 50 human-written scenarios with 921 generated scenarios, using a total of 971 scenarios.
For each scenario, only one dialogue is generated to ensure diversity.
Dialogues follow a turn distribution of 70\% with 2 turns, 20\% with 3 turns, and 10\% with 4 turns, yielding datasets with an average of 2.4 turns and 5,346 tokens.
Additionally, we generate 972 examples for the Summary LLM and 2,484 examples for the QA LLM.
This process takes approximately 2 hours and costs a total of \$72.76.
For ablation studies, all conditions remain identical to the original experiments, except that we use \texttt{gpt-4o-mini-2024-07-18} as the backbone to generate dialogue data cost-effectively across various experimental settings.
In this case, generating a similar amount of data to the GPT-4o experiment consumes approximately \$5.3.

\paragraph{Model Fine-Tuning}
The dialogue, summary, and QA datasets generated earlier are all used to fine-tune a single LLM.
We employ Supervised Fine-Tuning (SFT) to simultaneously train three tasks: language agent, summary LLM, and QA LLM.
When utilizing the model as the Tox-chat backbone, we use the same model for all three tasks to optimize LLM inference efficiency.
The training process utilizes Huggingface Transformers~\citep{wolf-etal-2020-transformers} and Accelerate~\citep{accelerate} libraries.
We optimize the model using the AdamW optimizer~\citep{loshchilov2018decoupled} at a learning rate of $3\times10^{-5}$.
We employ a maximum token length of 8,196 with a batch size of 16, and training proceeds over 3 epochs.
All other training hyper-parameters follow the default settings of Huggingface Trainer.
For Llama 3.1 8B, we utilize Fully-Sharded Data Parallel~\citep{zhao2023pytorchfsdpexperiencesscaling} on four A100-80GB GPUs, with the fine-tuning process taking approximately 4 hours and 20 minutes.

\begin{table*}[th]
\centering
\resizebox{\textwidth}{!}{%
\begin{tabular}{l|l}
\toprule
\multicolumn{2}{l}{\textbf{Question: Please tell me if there are any cases of exposure to hydroquinone.}}\\
\midrule
\textbf{Model} & \textbf{Response} \\
\midrule
Llama-8B & \begin{tabular}[l]{@{}l@{}} 
Hydroquinone is a chemical substance mainly used in photo development, cosmetics, and paints. 
\\ However, hydroquinone is a toxic substance, that can pose health risks with long-term exposure.
\\ Here are a few examples of cases where people have been exposed to hydroquinone.
\\1. \textbf{Cosmetic use}: Hydroquinone is a chemical used in cosmetics, primarily to lighten dark areas of skin. 
\\\quad However, with prolonged use, it can cause \fontblue{skin irritation or lead to skin cancer}. 
\\\quad The US FDA restricted hydroquinone in cosmetics in 2006.
\\2. \textbf{Photo development}: Hydroquinone is a chemical used in photo development to brighten photographs. 
\\\quad However, with long-term exposure, photo developers may experience \fontblue{skin irritation or respiratory problems}.
\\3. \textbf{Paint}: Hydroquinone is a chemical used in paint that makes the paint brighter.
\\\quad However, when exposed for long periods, paint workers may experience \fontblue{skin irritation or respiratory problems}.
\\\textit{⋯ (omitted)} \\
\end{tabular}
\\
\midrule
Toxchat-8B & \begin{tabular}[l]{@{}l@{}}  


Here are the documented cases of hydroquinone exposure:\\
1. \textbf{Occupational exposure case}: \fontpurple{A 43-year-old male suffered from myelodysplastic syndrome} 
\\\quad \fontpurple{and acute myeloid leukemia} after being exposed to hydroquinone for 16 years, 
\\\quad with chromosomal abnormalities observed [[Toxicity Information, Hydroquinone, Human Effects-Case Reports]].
\\
2. \textbf{Gastrointestinal disorders}: \fontpurple{544 people aboard a large US Navy vessel experienced symptoms}
\\\quad \fontpurple{such as nausea, vomiting, and abdominal pain}, which were determined to be caused by
\\\quad hydroquinone contamination [[Toxicity Information, Hydroquinone, Human Effects-Case Reports]].
\\
3. \textbf{Fatal case}: \fontpurple{A fatality occurred after ingestion of photo developer solution containing hydroquinone}, 
\\\quad with hydroquinone concentrations measured at 3.4 μg/mL in urine, 0.5 μg/g in liver, and 0.2 μg/g in kidney 
\\\quad[[Toxicity Information, Hydroquinone, Human Effects-Case Reports]].
\\\textit{⋯ (omitted)}
\end{tabular}
\\
\bottomrule
\end{tabular}
}
\caption{
Question and response example where Tox-chat-8B loses to Llama-8B in preference evaluation.
While Llama-8B presented \fontblue{general and redundant content}, Tox-chat-8B provided \fontpurple{specific and diverse cases} based on the database.
We have translated the original Korean text to English and omitted supplementary information.
}
\label{tab:qualitative-examples}
\end{table*}

\section{Qualitative Results}
\label{sec:qual-results}

As shown in Table~\ref{tab:qualitative-examples}, the DB-based answers generated by Tox-chat tend to be more diverse and factually specific
Overall, these results strongly support the effectiveness of our method.

\section{Human Agreement on LLM-as-a-judge}
\label{sec:human_agreement}
To validate the LLM-as-a-judge, we conduct a user study.
For 50 test examples and instructions, three human annotators judge preference and DB consistency.
The final human decision is determined by a majority vote.
Inter-annoator agreement is measured between human evaluators and the evaluation of LLM judge, using Kendall Tau ($\tau$) \citep{kendall1938new}. 
For the DB consistency, the $\tau$ is 0.315 
and Fleiss' Kappa score between annotators is 0.525.
For the preference, the $\tau$ is 0.337 
and the Kappa score is 0.358.
These scores reveal that the LLM evaluation is in moderate agreement with human evaluators.

\section{Detailed User Study Analysis}
\label{sec:appendix-user-study}

\begin{figure}[t]
    \centering
    \includegraphics[width=0.95\linewidth]{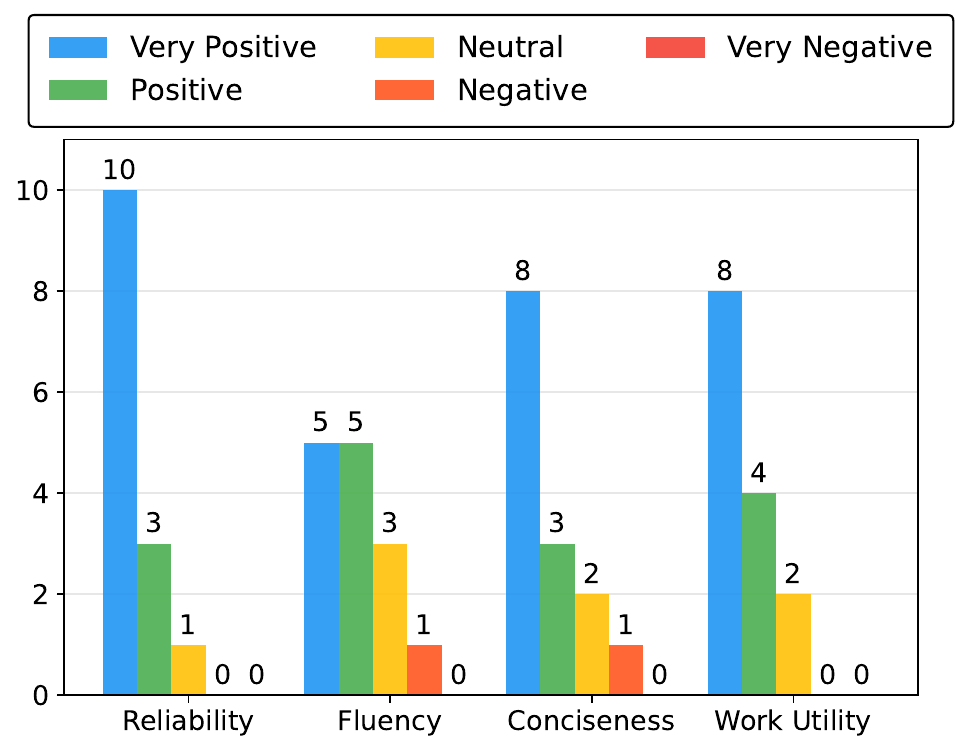}
    \caption{
    Experiment survey on Tox-chat compared with GPT-4o.
    }
    \label{fig:user-study-gpt4o}
\end{figure}

\begin{figure*}[ht]
    \centering
    \includegraphics[width=0.95\linewidth]{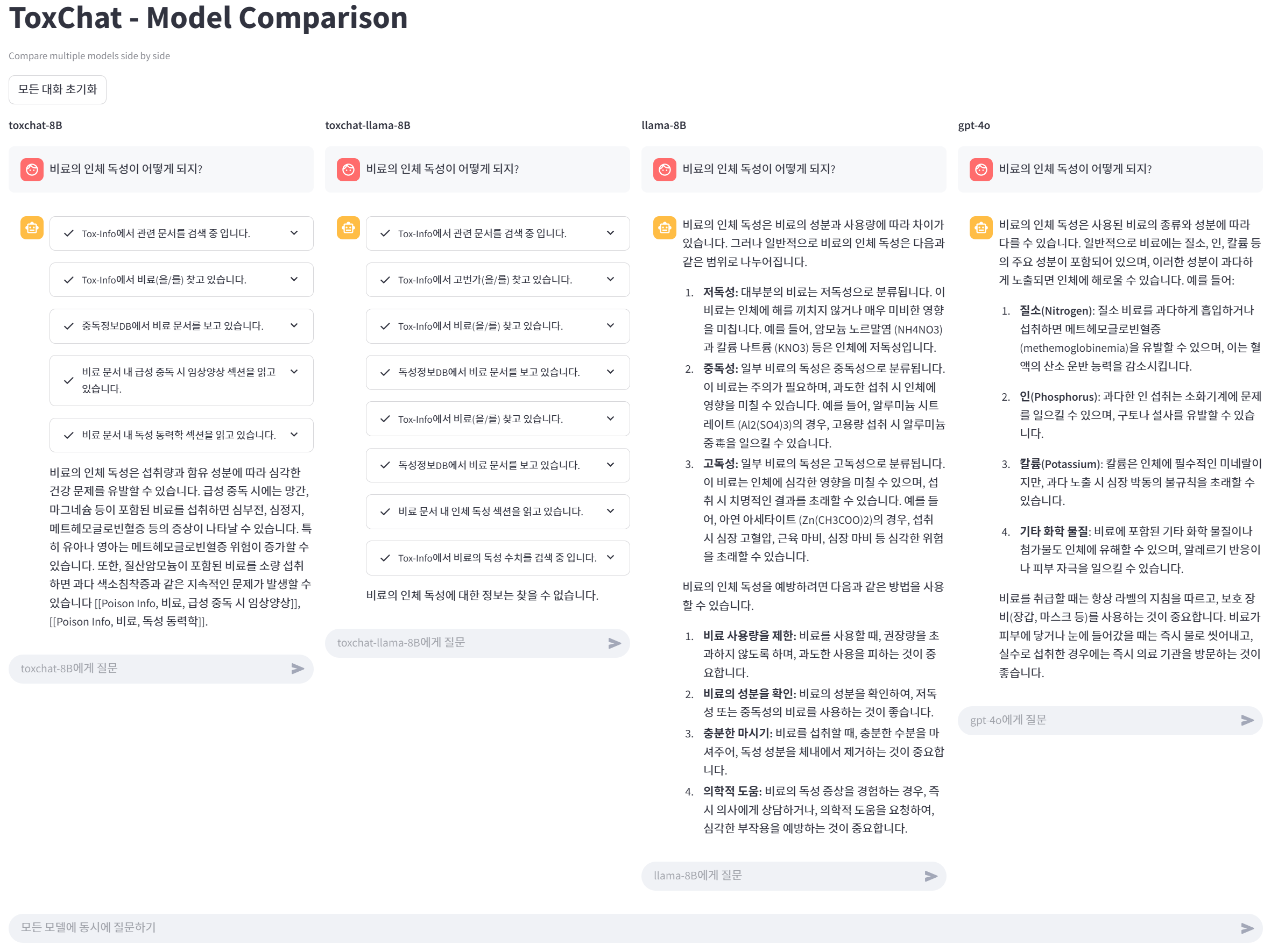}
    \caption{
    User interface used during the user study for model comparison. For models based on the Tox-chat architecture, the interface displays the tools invoked by the agent and the search results retrieved through tool usage.
    }
    \label{fig:model-comparison-UI}
\end{figure*}

Fig.~\ref{fig:model-comparison-UI} demonstrates the interface participants used during the user study. Users interacted directly with each model and subsequently compared their experiences.

Fig.~\ref{fig:user-study-gpt4o} presents 5-point scale survey results comparing the search experience between Tox-chat-8B and GPT-4o.  
These results demonstrate that Tox-chat was perceived as more reliable and useful for professional tasks than GPT-4o.

\section{Tox-Info DB Overview}
\label{sec:toxinfo_db_overview}
Tox-Info provides access to the following four key databases:
Tox-Info is a database system developed by the Korean Ministry of Food and Drug Safety, for the purpose of making information on chemicals used in food, medicines, and other products more easily accessible to both the general public and professionals.
Tox-info provides access to the following four key databases:

\paragraph{Chemical Info.}
This database contains information on chemicals used in products that directly interact with the human body, including food, medicines, and personal care items. It offers details on substances, usage, and toxicity information.

\paragraph{Poison Info.}
This database provides clinical toxicity and emergency treatment information for healthcare professionals, including doctors, nurses, and emergency treatment specialists, about toxic substances.

\paragraph{Cigarette Info.}
This database includes information on 93 harmful and potentially harmful constituents (HPHCs) in tobacco products and tobacco smoke, as designated by the U.S. Food and Drug Administration (FDA).

\paragraph{Carcinogen Info.}
This database offers information on carcinogenicity ratings, as classified by various international organizations, such as the International Agency for Research on Cancer (IARC), the National Toxicology Program (NTP), and the U.S. Environmental Protection Agency (EPA).

\paragraph{}
See Table~\ref{tab:full-document-example} for an English translation of one of the documents.
There are total 5,878 documents across Chemical Info DB and Poison Info DB, with over 80,000 sections. We measure the total number of tokens (using Llama-8B tokenizer) and report their statistics in Table~\ref{tab:toxinfo_stat}.

Note that while Cigarette Info DB and Carcinogen Info DB provide isolated search service, their documents are a subset of those in Chemical Info DB.
Additionally, the carcinogen filter search functionality is provided by Carcinogen Info DB, and the toxic dose search functionality is provided by Chemical Info DB.

\begin{table}[t]
\centering
\begin{tabular}{lll}
\toprule
\textbf{Statistic} & \textbf{Document} & \textbf{Section} \\ \hline
Min                & 70         & 22       \\ \hline
Max                & 57,179     & 12,354   \\ \hline
Mean               & 6,142.05   & 434.58   \\ \hline
Std. & 5,633.93   & 585.98   \\ \bottomrule
\end{tabular}
\caption{Token Length Statistics of Tox-Info DB, based on Llama-8B tokenizer.}
\label{tab:toxinfo_stat}
\end{table}

    

\section{Detailed Tool Description and Prompts}
\label{sec:prompts}
Table~\ref{tab:llm-tools-summary} describes the details of tools we use.
For dataset construction, the system and user prompts are described in Table~\ref{table:prompt-scenario-generation} and Table~\ref{table:prompt-user-persona}.
The full system prompt of our Tox-chat architecture are Table~\ref{table:prompt-toxchat-main}, and Table~\ref{table:prompt-toxchat-bm25-summary} and Table~\ref{table:prompt-toxchat-qa} show the module prompt.
Table~\ref{table:prompt-eval-db-consistency} and Table~\ref{table:prompt-eval-preference} are the instruction prompt for LLM-as-a-Judge.

\section{Generated Samples}
\label{sec:appendix-generated-data}
Table~\ref{table:example-generated-scenario} shows the generated scenario prompt when we build the training set, and Table~\ref{tab:train-dataset-example} shows an train set example. 

\begin{table*}[t]
\centering
\caption{Summary of tools in our Tox-chat architecture.}
\label{tab:llm-tools-summary}
\resizebox{\textwidth}{!}{%
\begin{tabular}{clll}
\toprule
\textbf{Tool Name} & \textbf{Functionality} & \textbf{Input(s)} & \textbf{Output} \\
\midrule
\texttt{BM25 Search} & \begin{tabular}[l]{@{}l@{}}Search ToxInfo using BM25 and summarize matched sections \\ with a separate LLM.\end{tabular} & Query string & Summarized text \\ 
\hline
\texttt{Keyword Search} & Performs a keyword search in ToxInfo. & Search keyword & \begin{tabular}[l]{@{}l@{}}List of highest matching \\ chemical names\end{tabular} \\
\hline
\texttt{Read General} & Retrieves abstract and table of contents from a chemical document. & Database name, \newline chemical name & \begin{tabular}[l]{@{}l@{}}Abstract \\ and table of contents\end{tabular} \\
\hline
\texttt{QA Specific} & \begin{tabular}[l]{@{}l@{}} Answers a question based on a specific section of a chemical document. \\ Returns the whole section if less than 200 characters\end{tabular} & \begin{tabular}[l]{@{}l@{}}Database name, \newline chemical name, \\ \newline section name, \newline question string \end{tabular}& \begin{tabular}[l]{@{}l@{}}Answer to the question \\ or the entire section \end{tabular}\\
\hline
\begin{tabular}[c]{@{}c@{}}\texttt{Carcinogen} \\  \texttt{Filter Search}\end{tabular} & \begin{tabular}[l]{@{}l@{}} Performs a keyword search in ToxInfo with optional carcinogenicity  \\ filters. Multiple filters can be applied, which are combined with \\ either 'AND' or 'OR' operator.\end{tabular}  & \begin{tabular}[l]{@{}l@{}} Keyword, \newline carcinogen filters \\ \newline (IARC/NTP/USEPA), \newline AND/OR flag   \end{tabular}& List of filtered chemical names \\
\hline
\begin{tabular}[c]{@{}c@{}}\texttt{Toxic Dose} \\  \texttt{Search}\end{tabular} & \begin{tabular}[l]{@{}l@{}} Retrieves toxic dose values like LD50, NOAEL, etc. for a chemical. \\ Structured info separated by equal signs; specific to dose data \end{tabular}& Chemical name & List of dose entries \\
\bottomrule
\end{tabular}%
}
\end{table*}

\begin{table*}
\centering
\caption{Generated scenario examples}
\resizebox{.9\textwidth}{!}{%
\begin{tabular}{l}
\toprule
\textbf{\textit{Example 1}} \\
\midrule
"persona":"50대 농부",
\\\qquad\qquad\quad \fontgray{A farmer in his 50s}
\\"situation":"농약 사용 시 안전성을 확인하려 함",
\\\qquad\qquad\quad \fontgray{Trying to confirm the safety when using pesticides}
\\"intention":"사용 중인 농약의 독성을 알고 싶음",
\\\qquad\qquad\quad \fontgray{Wants to know about the toxicity of the pesticide being used}
\\"question":"내가 쓰는 농약에 디아지논이 들어 있는데, 이게 사람한테 얼마나 위험한가요?"
\\\qquad\qquad\quad \fontgray{The pesticide I'm using contains diazinon. How dangerous is this to humans?}
\\
\midrule
\textbf{\textit{Example 2}} \\
\midrule
"persona": "화학 회사에 다니는 40대 연구원"
\\\qquad\qquad\quad \fontgray{A researcher in his 40s working at a chemical company}
\\"situation":"새로운 화학물질을 개발 중",
\\\qquad\qquad\quad \fontgray{Developing a new chemical substance}
\\"intention":"개발 중인 화학물질의 독성 평가를 하고 싶음",
\\\qquad\qquad\quad \fontgray{Wants to evaluate the toxicity of the chemical substance under development}
\\"question":"N-카보시메틸이미노비스(에틸렌니트로)테트라(아세트산)의 독성 정보를 알려줄 수 있어?"
\\\qquad\qquad\quad \fontgray{Can you provide toxicity information about N-carboxymethyliminobis(ethylenenitro)tetra(acetic acid)?}
\\
\midrule
\textbf{\textit{Example 3}} \\
\midrule
"persona": "20대 의학과 학생"
\\\qquad\qquad\quad \fontgray{A medical student in their 20s}
\\"situation":"독성학 과제 준비 중",
\\\qquad\qquad\quad \fontgray{Preparing for a toxicology assignment}
\\"intention":"부프레노르핀의 약리학적 독성을 조사하고자 함",
\\\qquad\qquad\quad \fontgray{Wants to research the pharmacological toxicity of buprenorphine}
\\"question":"부프레노르핀의 약리학적 독성을 간단히 설명해주세요."
\\\qquad\qquad\quad \fontgray{Please briefly explain the pharmacological toxicity of buprenorphine.}
\\
\midrule
\textbf{\textit{Example 4}} \\
\midrule
"persona": "환경문제에 관심이 많은 30대 여성"
\\\qquad\qquad\quad \fontgray{A woman in her 30s who is interested in environmental issues}
\\"situation":"최근에 수질오염 관련 다큐멘터리를 시청함",
\\\qquad\qquad\quad \fontgray{Recently watched a documentary about water pollution}
\\"intention":"도모이산이라는 물질이 언급되었는데, 궁금증이 생김",
\\\qquad\qquad\quad \fontgray{Became curious about a substance called domoic acid that was mentioned}
\\"question":"도모이산이라는 물질이 수질오염과 관련 있다고 하는데, 이게 어떤 독성을 가지고 있어?"
\\\qquad\qquad\quad \fontgray{I heard that a substance called domoic acid is related to water pollution. What kind of toxicity does it have?} 
\\
\bottomrule
\end{tabular}
}
\label{table:example-generated-scenario}
\end{table*}

\begin{table*}[htbp]
\centering
\caption{Tox-chat train dataset example. For readability, we omit the details (...) of documents.}
\resizebox{\textwidth}{!}{%
\begin{tabular}{cll}
\toprule
\textbf{Step} & \textbf{Component} & \textbf{Content} \\
\midrule
\textbf{1} & \textbf{User Query} & 
\begin{tabular}[l]{@{}l@{}} 
인간이 베릴륨에 노출되면 어떤 증상이 나타나? \\
\fontgray{What symptoms appear when humans are exposed to beryllium?}
\end{tabular}
\\
\midrule
2 & Agent Tool Call & 
\begin{tabular}[l]{@{}l@{}} 
\texttt{tool\_call: BM25 Search} \\
\texttt{arguments: \{"query":"인간이 베릴륨에 노출되면 어떤 증상이 나타나?"\}} 
\end{tabular}
\\
\midrule
3 & Tool Response & \begin{tabular}[l]{@{}l@{}} 
인간이 베릴륨에 노출될 경우 다양한 증상이 나타날 수 있다. 급성 노출 시에는 화학적 폐렴, 기관지염, 결막염, \\ 
피부염 등이 발생하며, 심한 경우 치명적일 수 있다. 만성 노출은 폐육아종증, 체중 감소, 호흡 부전, 간 및 비장 비대 \\
등의 증상을 유발할 수 있다... \\
\fontgray{When humans are exposed to beryllium, various symptoms can appear. In acute exposure, chemical pneumonia, bronchitis,}  \\
\fontgray{conjunctivitis, dermatitis, etc. can occur, and in severe cases, it can be fatal. Chronic exposure can cause pulmonary} \\
\fontgray{granulomatosis, weight loss, respiratory failure, liver and spleen enlargement...}
\end{tabular}
\\
\midrule
4 & Agent Tool Call & 
\begin{tabular}[l]{@{}l@{}} 
\texttt{tool\_call: Keyword Search} \\
\texttt{arguments: \{"keyword":"베릴륨"\}} \fontgray{Beryllium}
\end{tabular}
\\
\midrule
5 & Tool Response &\begin{tabular}[l]{@{}l@{}} 
\# Search results for 베릴륨 in ToxInfo database. \\
\#\# Chemical Info Search Results (From most relevant to least): \\
1. 베릴륨 (Beryllium, elemental) \\
2. 산화베릴륨 (Beryllium oxide) \\
3. 인산베릴륨 (Beryllium phosphate)...
\end{tabular}
\\
\midrule
6 & Agent Tool Call &
\begin{tabular}[l]{@{}l@{}} 
\texttt{tool\_call: Read General} \\
\texttt{arguments: \{"db\_name":"Chemical Info","chemical\_name":"베릴륨"\}} 
\end{tabular} 
\\
\midrule
7 & Tool Response & \begin{tabular}[l]{@{}l@{}} \# General Information for 베릴륨 in Chemical Info database \\
\#\# Abstract \\
베릴륨 화합물(Beryllium compound)은 육각형 결정의 형태 흰색을 나타내며, 냄새가 나지 않는 금속이다. \\ 
주로 구리 베릴륨 합금에서 합금원소로 사용되며...\\
\fontgray{Beryllium compounds appear as white hexagonal crystals and are odorless metals. They are mainly used as an alloying } \\ 
\fontgray{element in copper beryllium alloys...}
\end{tabular}\\
\midrule
8 & Agent Tool Call & 
\begin{tabular}[l]{@{}l@{}} 
\texttt{tool\_call: QA Specific} \\
\texttt{arguments: \{"db\_name":"Chemical Info","chemical\_name":"베릴륨","section\_name":"인체영향-증상",} \\
\texttt{"question":"인간이 베릴륨에 노출되면 어떤 증상이 나타나?"\}} \\
\fontgray{arguments: \{"db\_name":"Chemical Info","chemical\_name":"Beryllium","section\_name":"Human Health Effects-Symptoms",}\\
\fontgray{"question":"What symptoms appear when humans are exposed to beryllium?"\}}
\end{tabular}
\\
\midrule
9 & Tool Response &  \begin{tabular}[l]{@{}l@{}} 
인간이 베릴륨에 노출되면 다음과 같은 증상이 나타날 수 있습니다: \\
1. \textbf{흡입 노출}: 급성 폐렴, 흉통, 기관지 경련, 발열, 호흡 곤란, 청색증, 기침... \\
2. \textbf{피부 접촉}: 먼지와의 접촉으로 급성 피부염이 발생하며...\\
3. \textbf{눈 접촉}: 먼지 오염으로 인해 급성 결막염이 발생하며...\\
\fontgray{When humans are exposed to beryllium, the following symptoms may appear:} \\
\fontgray{1. \textbf{Inhalation Exposure}: Acute pneumonia, chest pain, bronchospasm, fever...}\\
\fontgray{2. \textbf{Skin Contact}: Acute dermatitis can occur from contact with dust...} \\
\fontgray{3. \textbf{Eye Contact}: Acute conjunctivitis can occur due to dust contamination...}
\end{tabular}
\\
\midrule
\textbf{10} & \textbf{\fontpurple{Agent Response}} &  \begin{tabular}[l]{@{}l@{}}
인간이 베릴륨에 노출되면 다음과 같은 증상이 나타날 수 있습니다:\\
1. \textbf{흡입 노출}: 급성 폐렴, 흉통, 기관지 경련, 발열, 호흡 곤란, 기침 등이 발생하며, 만성 노출 시 체중 감소, 심장 비대, \\
\quad 간 및 비장 비대 등이 나타날 수 있습니다.\\
2. \textbf{피부 접촉}: 급성 피부염, 홍반, 수포성 발진 등이 발생할 수 있습니다.\\
3. \textbf{눈 접촉}: 급성 결막염, 결막 충혈, 광선 공포증 등이 보고되었습니다...\\
\fontgray{When humans are exposed to beryllium, the following symptoms can appear:} \\
\fontgray{1. \textbf{Inhalation Exposure}: Acute pneumonia, chest pain, bronchospasm...} \\
\fontgray{2. \textbf{Skin Contact}: Acute dermatitis, erythema, vesicular rash... }\\
\fontgray{3. \textbf{Eye Contact}: Acute conjunctivitis, conjunctival hyperemia... }
\end{tabular}
\\
\midrule
\textbf{11} & \begin{tabular}[l]{@{}l@{}} \textbf{User Follow-up} \\ \textbf{Query} \end{tabular}& 
\begin{tabular}[l]{@{}l@{}} 
베릴륨에 노출되는 주요 경로는 어떤 방식으로 이루어지나요? 예를 들어, 주로 작업 환경에서 발생하는 건가요? \\
\fontgray{What are the main routes of exposure to beryllium? For example, does it mainly occur in work environments?}
\end{tabular}
\\
\midrule
12 & Agent Tool Call &
\begin{tabular}[l]{@{}l@{}}
\texttt{tool\_call: QA Specific} \\
\texttt{arguments: \{"db\_name":"Chemical Info","chemical\_name":"베릴륨","section\_name":"인체영향-사례보고",}\\ 
\texttt{"question":"베릴륨에 노출되는 주요 경로는 무엇인가요? 작업 환경에서 주로 발생하나요?"\}} \\
\fontgray{arguments: \{"db\_name":"Chemical Info","chemical\_name":"Beryllium","section\_name":"Human Health Effects-Case Reports",} \\
\fontgray{"question":"What are the main routes of exposure to beryllium? Does it primarily occur in work environments?"\}}
\end{tabular}
\\
\midrule
13 & Tool Response & 
\begin{tabular}[l]{@{}l@{}} 베릴륨에 노출되는 주요 경로는 작업 환경에서 발생하는 경우가 많습니다. 베릴륨은 주조 작업, 주물, 전단, 압연, 절단, \\
 용접, 기계 가공, 샌딩, 연마, 조립 및 화학 분석 작업 중에 노출될 수 있습니다... \\
\fontgray{The main routes of exposure to beryllium often occur in work environments. Beryllium exposure can happen } \\
\fontgray{during casting operations, molding, shearing, rolling, cutting, welding, machining, sanding, grinding, assembly,} \\
\fontgray{ and chemical analysis work... }
\end{tabular}\\
\midrule
\textbf{14} & \textbf{\fontpurple{Agent Response}}& 
\begin{tabular}[l]{@{}l@{}} 
베릴륨에 노출되는 주요 경로는 주로 작업 환경에서 발생합니다. 주조 작업, 용접, 기계 가공, 연마 등 다양한 산업 활동 중에 \\
베릴륨에 노출될 수 있습니다. 또한, 작업복을 통해 집 먼지로 옮겨지거나, 베릴륨 공장 근처 대기 오염으로 인해 작업장 \\ 
외부에서도 노출 사례가 보고된 바 있습니다... \\
\fontgray{The main routes of exposure to beryllium primarily occur in work environments. Exposure can happen during} \\ 
\fontgray{various industrial activities such as casting operations, welding, machining, grinding, and more.} \\
\fontgray{Additionally, exposure cases have been reported outside the workplace} \\
\fontgray{through work clothes transferring to house dust, or through air pollution near beryllium factories...}
\end{tabular}\\
\bottomrule
\end{tabular}
}
\label{tab:train-dataset-example}
\end{table*}

\begin{table*}[]
\centering
\caption{Full example of Table~\ref{tab:qualitative-examples}. Naive backbone models generate \fontblue{general responses}, but Tox-chat models refer \fontpurple{detailed domain-specific informaton.}}
\renewcommand{\arraystretch}{0.97}
\resizebox{\textwidth}{!}{%
\begin{tabular}{l|l}
\toprule
\multicolumn{2}{l}{\textbf{Question: Please tell me if there are any cases of exposure to hydroquinone.}}\\
\midrule
\textbf{Model} & \textbf{Response} \\
\midrule
GPT-4o & \begin{tabular}[l]{@{}l@{}}
Hydroquinone is a chemical mainly used in skin whitening creams that has the effect of inhibiting melanin production \\
to brighten skin. However, this substance is regulated in many countries due to side effects and safety concerns. \\
Cases of exposure to hydroquinone are mainly associated with the following side effects.
\\
1. \textbf{Skin irritation and allergic reactions}: Symptoms such as \fontblue{skin rash, itching, and erythema} may occur \\
after using products containing hydroquinone.\\
2. \textbf{Ochronosis}: Long-term use can lead to a condition called ochronosis, where the skin turns bluish or darkens. 
\\\quad This mainly occurs when high concentrations of hydroquinone are used for extended periods.\\
3. \textbf{Carcinogenicity controversy}: Some studies suggest that hydroquinone may have carcinogenic potential, 
\\\quad although clear evidence in humans is lacking. Nevertheless, due to this possibility, some countries prohibit or 
\\\quad restrict the use of hydroquinone. \\
4. \textbf{Eye and respiratory irritation}: Exposure to hydroquinone powder or vapor can \fontblue{irritate the eyes and respiratory system}.\\
For these reasons, it is important to follow professional guidance or strictly adhere to product usage instructions \\
when using hydroquinone.  It is also advisable to consult with a dermatologist when using products containing hydroquinone.
\end{tabular}
\\
\midrule
\textbf{Tox-chat [GPT-4o]} & \begin{tabular}[l]{@{}l@{}}
Cases of exposure to hydroquinone are as follows:
\\ 1. \textbf{Occupational exposure}: \fontpurple{A 43-year-old male was diagnosed with myelodysplastic syndrome and acute myeloid leukemia} 
\\\quad after 16 years of occupational exposure to radiographic developer fluid, with chromosomal abnormalities observed 
\\\quad [[Toxicity information, Hydroquinone, Human effects-case reports]].
\\ 2. \textbf{Acute gastrointestinal disease}: \fontpurple{544 crew members of a US Navy vessel experienced acute gastrointestinal disease }
\\\quad \fontpurple{including nausea, vomiting, and abdominal pain} due to a cold water system contaminated with hydroquinone 
\\\quad [[Toxicity information, Hydroquinone, Human effects-case reports]].
\\ 3. \textbf{Fatal case}s: Fatal cases occurred due to ingestion of hydroquinone-containing photographic developer fluid, 
\\\quad and autopsy showed \fontpurple{hydroquinone concentrations} in \fontpurple{urine, liver, and kidneys}
\\\quad [[Toxicity information, Hydroquinone, Human effects-case reports]].
\\ 4. \textbf{Respiratory and skin effects}: Workers exposed to hydroquinone mixtures had higher
\\ \fontpurple{prevalence of respiratory symptoms and eczema than the control group}, 
\\\quad and lower lung function values [[Toxicity information, Hydroquinone, Human effects-epidemiological studies]].
\\ 5. \textbf{Cancer incidence study}: A cohort study of workers exposed to hydroquinone at a Tennessee factory found \\
\fontpurple{cancer incidence rates similar to or lower than} \fontpurple{the general population} 
\\\quad [[Toxicity information, Hydroquinone, Human effects-epidemiological studies]].
\\ 6. \textbf{Increased malignant melanoma}: Some Danish lithographers were regularly exposed to hydroquinone, 
\\\quad and the \fontpurple{incidence of malignant melanoma increased}
\\\quad [[Toxicity information, Hydroquinone, Human effects-epidemiological studies]].
\\ These cases show various effects related to the toxicity of hydroquinone.
\end{tabular}
\\
\midrule
Llama-8B & \begin{tabular}[l]{@{}l@{}}
Hydroquinone is a chemical substance mainly used in photo development, cosmetics, and paints. However, hydroquinone is 
\\ a toxic substance, that can pose health risks  with long-term exposure. Here are a few examples of cases 
\\ where people have been exposed to hydroquinone.
\\ 1. \textbf{Cosmetic use}: Hydroquinone is a chemical used in cosmetics, primarily to lighten dark areas of skin. 
\\\quad However, with prolonged use, it can cause \fontblue{skin irritation or lead to skin cancer}. 
\\\quad The US FDA restricted hydroquinone in cosmetics in 2006.
\\ 2. \textbf{Photo development}: Hydroquinone is a chemical used in photo development to brighten photographs.
\\\quad However, with long-term exposure, photo developers 
\\\quad may experience \fontblue{skin irritation or respiratory problems}.
\\ 3. \textbf{Paint}: Hydroquinone is a chemical used in paint that makes the paint brighter. 
\\\quad However, when exposed for long periods, paint workers may experience 
\\\quad \fontblue{skin irritation or respiratory problems}.
\\ 4. \textbf{Industrial exposure}: Hydroquinone can also pose \fontblue{health risks} with industrial exposure. 
\\\quad For example, factory workers who produce or process hydroquinone
\\\quad may \fontblue{experience skin irritation or respiratory problems} with prolonged exposure.\\
One of the cases of exposure to hydroquinone is a 2019 incident in the US where photo developers were exposed to hydroquinone 
\\ and experienced skin irritation. In this incident, photo developers in the US  used hydroquinone for a long period, \\ 
experienced skin irritation, and were eventually diagnosed with skin cancer. Cases of exposure to hydroquinone can occur \\
in various industries and usage areas. Therefore, people who use or are exposed to hydroquinone should follow safe usage  
\\ methods or use protective equipment to protect their health.
\end{tabular}
\\
\midrule
\textbf{Tox-chat [Llama-8B]} & \begin{tabular}[l]{@{}l@{}}
There are cases of exposure to hydroquinone. \fontpurple{A 43-year-old male was diagnosed with myelodysplastic syndrome and} \\
\fontpurple{acute myeloid leukemia} after occupational exposure to hydroquinone in radiographic developer fluid for 16 years. \\
Gastrointestinal illness characterized by \fontpurple{acute symptoms including nausea, vomiting, abdominal pain,} \\
\fontpurple{and diarrhea occurred in 544 crew members} aboard a large US Navy vessel, which was found to be caused \\
by contamination of the cold water system with hydroquinone originating from an automatic photo processing machine \\
installed on the vessel.
\end{tabular}
\\
\midrule
\textbf{Tox-chat-8B} & \begin{tabular}[l]{@{}l@{}}  
Here are the documented cases of hydroquinone exposure:\\
1.\textbf{Occupational exposure case}: \fontpurple{A 43-year-old male suffered from myelodysplastic syndrome and acute myeloid leukemia} 
\\\quad after being exposed to hydroquinone for 16 years, with chromosomal abnormalities observed 
\\\quad {[}{[}Toxicity Information, Hydroquinone, Human Effects-Case Reports{]}{]}.
\\
2. \textbf{Gastrointestinal disorders}: \fontpurple{544 people aboard a large US Navy vessel experienced symptoms such as nausea, }
\\\quad \fontpurple{vomiting, and abdominal pain}, which were determined to be caused by hydroquinone contamination 
\\\quad {[}{[}Toxicity Information, Hydroquinone, Human Effects-Case Reports{]}{]}.
\\
3. \textbf{Fatal case}: \fontpurple{A fatality occurred after ingestion of photo developer solution containing hydroquinone},
\\\quad with hydroquinone concentrations measured at 3.4 μg/mL in urine, 0.5 μg/g in liver, and 0.2 μg/g in kidney 
\\\quad {[}{[}Toxicity Information, Hydroquinone, Human Effects-Case Reports{]}{]}.\\
Additionally, evidence of hydroquinone's carcinogenicity in humans is inadequate, and in animal experiments,
\\ it was evaluated that carcinogenicity classification is not possible  due to limited evidence.
\end{tabular}
\\
\bottomrule
\end{tabular}
}
\label{tab:full-qualitative-examples}
\end{table*}

\begin{table*}
\centering
\caption{Prompts for scenario generation. \fontgray{Gray text} is a reference translation of Korean into English, and was not actually included in the model input.
\fontpurple{Purple text} is a variable that changes for each example.
}
\resizebox{\textwidth}{!}{%
\begin{tabular}{l}
    \toprule
\textit{\textbf{System prompt}} \\
\midrule
You are writer for a Toxicity-info chatbot usage scenarios.\\
Create a scenario that are likely to be asked.\\
Often, the scenario is related to a certain chemical or product.\\
Be as creative and diverse as possible.\\
Do not use a similar material that's given in the previous response.\\
Do not create duplicate scenarios that share similar material or idea.\\
Write in Korean.\\
\\
Here are some important rules you must follow when writing the scenario, especially the Question:\\
\quad1. If you are putting a chemical into a scenario, use a specific name.\\
\quad2. If you are putting a product into a scenario, refer to a certain product type.\\
\quad3. Do not create vague questions that are impossible to answer without any follow-up question.\\
\quad4. You must give enough context and information in the question, so that they can be answered by the chatbot in one go.\\
\quad5. Here are some chemicals you may use to create your scenarios on. Use them only when they naturally fit to the scenario.\\
\qquad-\textcolor{fontpurple}{\{EXAMPLE\_CHEMICALS\}} \\
\midrule

\textit{\textbf{User prompt}} \\
\midrule
\{"scenarios": {[} \\
\quad \{
\\\qquad"persona":"20대 상하차 직원",
\\\qquad\qquad\qquad \fontgray{A warehouse worker in his 20s}
\\\qquad"situation":"트럭에 둘러싸여서 일하다 보니, 매연을 하루종일 맡고 있는게 걱정된다.",
\\\qquad\qquad\qquad \fontgray{Working surrounded by trucks, I'm worried about inhaling exhaust fumes all day.}
\\\qquad"intention":"독성 챗봇에게 매연의 위험성에 대해 물어보고자 한다.",
\\\qquad\qquad\qquad \fontgray{User wants to ask a toxic chatbot about the dangers of exhaust fumes.}
\\\qquad"question":"매연을 하루종일 맡고 있는데, 이게 얼마나 안좋은거야"
\\\qquad\qquad\qquad \fontgray{I'm inhaling exhaust fumes all day, how bad is this for me?}
\\\quad\}, \\
\quad \{
\\\qquad"persona":"40대 남성",
\\\qquad\qquad\qquad \fontgray{A male around age 40}
\\\qquad"situation":"작업장에서 암모니아를 많이 사용하고있는 상황",
\\\qquad\qquad\qquad \fontgray{In a workplace where ammonia is frequently used}
\\\qquad"intention":"작업장에 암모니아를 사용하는데 인체에 많이 해로운지 궁금",
\\\qquad\qquad\qquad \fontgray{Curious about how harmful ammonia is to the human body when used in the workplace}
\\\qquad"question":"작업장에 암모니아를 사용하는데 인체에 많이 해로울까요?"
\\\qquad\qquad\qquad \fontgray{We use ammonia in the workplace, is it very harmful to humans?}
\\\quad\}, 
\\\quad ... 
\\\quad\{
\\\qquad"persona":"논문 제출을 앞둔 20대 공대 대학원생",
\\\qquad\qquad\qquad \fontgray{Engineering graduate student in their 20s, about to submit a paper}
\\\qquad"situation":"나는 제로콜라를 하루에 1.5L 씩 마시며, 최근 화장실에 너무 자주 가는 것 같다.",
\\\qquad\qquad\qquad \fontgray{I drink 1.5L of zero-calorie cola daily, and recently I seem to be going to the bathroom too frequently.}
\\\qquad"intention":"독성 챗봇에게 이렇게 콜라를 많이 마셔도 되는지 물어보고자 한다.",
\\\qquad\qquad\qquad \fontgray{I want to ask a toxic chatbot if it's okay to drink this much cola.}
\\\qquad"question":"콜라 많이 마시면 뭐가 안 좋아?" 
\\\qquad\qquad\qquad \fontgray{What's bad about drinking a lot of cola?}
\\\quad\} \\
{]}\} \\
Generate other 10 scenarios in the JSON list based on the given examples.\\
\bottomrule
\end{tabular}
}
\label{table:prompt-scenario-generation}
\end{table*}

\begin{table*}
\centering
\caption{User simulation prompt}
\resizebox{.8\textwidth}{!}{%
\begin{tabular}{l}
\toprule
\textbf{\textit{System prompt}} \\
\midrule
You are a method actor that is committed to act the following scenario.\\
You will be given who you are acting and what scenario you are in.\\
Based on the chat history, create the most natural follow up question (or response) possible.\\
The scenarios will be given in Korean, and you can only use Korean as well.\\
\\
=== Beginning of scenario ===\\
You are a human that asks questions to a chatbot.\\
The chatbot is a specialist in toxic chemicals.\\
Below are specific details:\\
\#\#\# Character\\
\textcolor{fontpurple}{\{CHARACTER\}}\\
\\
\#\#\# Situation\\
\textcolor{fontpurple}{\{SITUATION\}}\\
\\
\#\#\# Intention\\
\textcolor{fontpurple}{\{INTENTION\}}\\
=== End of scenario ===\\
\\
The chatting started with the following question:\\
\textcolor{fontpurple}{\{USER\_QUERY\}}\\
\midrule
\textit{\textbf{User prompt}} \\
\midrule
\fontpurple{\{AGENT\_RESPONSE\}}
\\
\bottomrule
\end{tabular}
}
\label{table:prompt-user-persona}
\end{table*}

\begin{table*}
\centering
\caption{Tox-chat main prompt. \fontgray{Gray text} is just for reference.}
\resizebox{\textwidth}{!}{%
\begin{tabular}{l}
    \toprule
    \textbf{\textit{System prompt}} \\
\midrule
You are a helpful assistant that answers toxicity related answers based on given database.\\
The database, named Tox-Info, is comprised of four smaller databases:\\
1. Chemical Info: Provides information on chemicals used in products that come into direct contact with the human body, 
\\\quad such as food, medicines, and other personal care products, including substance information, usage, and toxicity information. \\
2. Poison Info: Offers clinical toxicity information and emergency treatment information for toxic substances to  
\\\quad healthcare professionals, including doctors, nurses, and emergency treatment specialists. \\
3. Cigarette Info: Provides information on 93 harmful and potentially harmful constituents (HPHCs) in tobacco products and 
\\\quad tobacco smoke, as designated by the U.S. Food and Drug Administration (FDA). \\
4. Carcinogen Info: Provides information on carcinogenicity ratings classified by various international organizations and agencies, 
\\\quad including the International Agency for Research on Cancer (IARC), the National Toxicology Program (NTP), 
\\\quad and the U.S. Environmental Protection Agency (EPA). \\
\\
When the user asks about a certain chemical, product, or creature, follow this protocol.\\
1. Always start with `toxinfo\_bm25\_summarize' tool.\\
\quad - Use the user's full question as the search keyword. If the question is more than 2 sentences, summarize it into one.\\
\quad - Since the result is not always accurate, you must decide whether each of the matched texts indeed are what the user is asking for. \\
\quad - Do not jump onto the result right away. Follow the next steps.\\
2. Next, use the `toxinfo\_db\_search' tool to find the document of the chemical in question.\\
3. If `toxinfo\_db\_search' again, do not jump right into conclusion. Try it again with different keywords. 
\\\quad In doing so, follow the rules below:\\
\quad a. You can use your own knowledge to modify the search keyword into a similar/related substance. For example,\\
\quad \quad - $\text{주유소}_{{\text{\fontgray{gas station}}}}$ is related to $\text{벤젠}_{{\text{\fontgray{benzene}}}}$, $\text{톨루엔}_{{\text{\fontgray{toluene}}}}$, $\text{에틸베젠}_{{\text{\fontgray{ethylbenzene}}}}$, $\text{자일렌}_{{\text{\fontgray{xylene}}}}$, etc.\\
\quad \quad - The keyword $\text{살균}_{{\text{\fontgray{disinfection}}}}$ is related to $\text{에탄올}_{{\text{\fontgray{ethanol}}}}$, $\text{과산화수소}_{{\text{\fontgray{hydrogen peroxide}}}}$, $\text{차아염소산 소듐}_{{\text{\fontgray{sodium hypochlorite}}}}$, etc.\\
\quad \quad - The keyword $\text{말벌}_{{\text{\fontgray{hornet}}}}$ is related to $\text{벌}_{{\text{\fontgray{bee}}}}$.\\
\quad b. If the substance is comprised of multiple chemicals, use your own knowledge to make a list of those chemicals. For example,\\
\quad \quad - $\text{담배}_{{\text{\fontgray{tobacco/cigarette}}}}$ is mainly comprised of $\text{니코틴}_{{\text{\fontgray{nicotine}}}}$, $\text{타르}_{{\text{\fontgray{tar}}}}$, $\text{일산화탄소}_{{\text{\fontgray{carbon monoxide}}}}$, $\text{벤젠}_{{\text{\fontgray{benzene}}}}$, etc.\\
\quad \quad - $\text{살충제}_{{\text{\fontgray{pesticide}}}}$ is mainly comprised of $\text{파라티온}_{{\text{\fontgray{parathion}}}}$, $\text{말라티온}_{{\text{\fontgray{malathion}}}}$, $\text{퍼메트린}_{{\text{\fontgray{permethrin}}}}$, etc.\\
\quad \quad - $\text{콜라}_{{\text{\fontgray{cola}}}}$ is mainly comprised of $\text{설탕}_{{\text{\fontgray{sugar}}}}$, $\text{카페인}_{{\text{\fontgray{caffeine}}}}$, $\text{인산}_{{\text{\fontgray{phosphoric acid}}}}$, etc.\\
\quad \quad - $\text{락스}_{{\text{\fontgray{bleach}}}}$ is mainly comprised of $\text{차아염소산 소듐}_{{\text{\fontgray{sodium hypochlorite}}}}$.\\
\quad c. With the new list, search them again using the `toxinfo\_db\_search' tool.\\
\quad d. Search the word in either English or Korean, not both.\\
4. Read through the database to find the information the user is asking for.\\
\quad - Use `read\_general\_chemical\_info' to get the general info and section names.
\\ \quad- When using `chemical\_info\_qa', make sure you use the exact section names obtained from `read\_general\_chemical\_info'.
\\ \quad- To access more specific information, use `chemical\_info\_qa'. In this case, you may have to access multiple sections \\
\qquad before finding the information you need, especially when the user is asking multiple questions.
\\ \quad- Even if there is only one  question, accessing multiple sections is a good practice.
\\ \quad- You may also have to access multiple documents, if the user is asking about multiple chemicals. \\
5. Upon gathering enough information, answer the user's question based on the information in the database.
\\ \quad - Your answer must be thoroughly fact-checked by the content in the database. Do not add facts of your own that are \\ \qquad not included in the database.
\\ \quad - When stating facts and numbers, you must make sure the information is identical to that written in the database.
\\ \quad - Do not include any redundant information, or those that the user didn't ask for. Try to keep the answers light.
\\ \quad - If the text contains any reference, include it in your answer as well.
\\ \quad - Always answer in Korean. \\
6. When answering the question, make sure that you reference the information in the database. When doing so, \\ \quad use the format of [[Database name, Chemical name, Section name]]. 
\\ \quad - Reference the information only when they are from the database.
\\ \quad - Do not make reference of your own. \\
7. In a special case where the question is about toxic dose information, use the `get\_toxic\_dose\_info' tool.
\\ \quad- Toxic dose information include but not limited to: LD50, LC50, TD50, NOAEL, ...\\
8. In a special case where the question is about finding chemicals with certain carcinogen class, \\ \quad use the `toxinfo\_db\_search\_with\_carcinogen\_filter' tool. 
\\ \quad - You can search for chemicals that are classified by IARC, NTP, USEPA.
\\ \quad - IARC Class 1, NTP Class K, USEPA Class A are the classes for the most carcinogenic chemicals.\\
\midrule 
\textbf{\textit{User Prompt}} \\
\midrule 
\fontpurple{\{USER\_QUERY\}}\\
\bottomrule
\end{tabular}
}
\label{table:prompt-toxchat-main}
\end{table*}

\begin{table*}
\centering
\caption{Tox-chat BM25 summarization prompt}
\resizebox{\textwidth}{!}{%
\begin{tabular}{l}
    \toprule
    \textbf{\textit{System prompt}} \\
\midrule
Your job is to summarize any given text into less than 200 words. \\
The text is a result of a database search. You will be given the text, and the query that was used to retrieve the texts. \\
When summarizing, filter out those that are completely unrelated to the given query. \\
Don't be too strict; if the text weakly relates, try to include them as well. \\
However, do not add any additional information of your own. Only use the text that's given in the text. \\
You will be given the query first, and then the retrieved text. \\
Your answer must include two types of references: \\
\quad 1. Reference from the given context, which are often surrounded by parantheses "( )". If the information has this reference, 
\\\qquad include in your response as well. \\
\quad 2. Reference indicating the location within the database. When referencing, use the format of 
\\\qquad [[Database name, Chemical name, Section name]]. \\
Reference the information if and only if when they are from the database. \\
Do not make reference of your own.\\
Summarize in Korean.\\

\midrule 
\textbf{\textit{User Prompt}} \\
\midrule 
\#\# Query \\
\fontpurple{\{QUERY\}}\\
\\
\#\# Retrieved Text\\
\fontpurple{\{RETRIEVED\_SECTION\_1\}}\\
\\
\fontpurple{\{RETRIEVED\_SECTION\_2\}}\\
...\\
\fontpurple{\{RETRIEVED\_SECTION\_10\}}\\

\bottomrule
\end{tabular}
}
\label{table:prompt-toxchat-bm25-summary}
\end{table*}

\begin{table*}
\centering
\caption{Tox-chat QA prompt}
\resizebox{\textwidth}{!}{%
\begin{tabular}{l}
\toprule
\textbf{\textit{System prompt}} \\
\midrule
Read the following context and answer the question in Korean. \\
Wherever possible, your answer must be thoroughly fact-checked by the information provided in the context.\\
When stating facts (e.g., numbers, properties, etc.), they should be directly referenced from the context.\\
Your answer should be compact, but you still must faithfully answer the question(s).\\
Your answer must not be longer than the context.\\
If the question is impossible the answer from the context, you must state that there is no information in the database to answer the question.\\
Your answer must include two types of references:\\
\quad 1. Reference from the given context, which are often surrounded by parantheses "( )". If the information has this reference, 
\\\qquad include in your response as well.\\
\quad 2. Reference indicating the location within the database. When referencing, use the format of 
\\\qquad [[Database name, Chemical name, Section name]].\\
Reference the information if and only if when they are from the database.\\
Do not make reference of your own.\\
\\
\# Context \\
\fontpurple{\{CONTEXT\}} \\

\midrule
\textbf{\textit{User Prompt}} \\
\midrule 
\fontpurple{\{
QUESTION\_FROM\_AGENT
\}
}
\\
\bottomrule
\end{tabular}
}
\label{table:prompt-toxchat-qa}
\end{table*}

\begin{table*}
\centering
\caption{LLM-as-a-judge prompt on DB consistency}
\resizebox{\textwidth}{!}{%
\begin{tabular}{l}
    \toprule
\textbf{\textit{System prompt}} \\
\midrule
Please act as an impartial judge and evaluate whether the response provided by an AI assistant is factually grounded. \\ 
You will be given a list of documents, each of which contain properties and dangers of a certain chemical. \\
Carefully go through the AI assistant's response. For any chemical fact you encounter, check whether it is \\
verifiable by the provided chemical documents. \\
Be as objective as possible. Briefly summarize the AI assistant's response, then decide your verdict on whether the response is fact-checked \\ by the documents. \\
Do not evaluate the fluency or helpfulness of the response. \\
Do not evaluate whether the question is answered or not. \\
Do not make your verdict based on whether the response contains every information in the chemical documents. \\
You will be given an excessive amount of broad information; the response does not need to cover every single chemical document to be \\ considered fact-checked. \\
In a special case where there are no documents provided, make your verdict based on the following rules:\\
\quad 1. If you are certain that the response is objectively wrong, make your decision based on your own knowledge.\\
\quad 2. If you are unsure whether the facts are verifiable, return a special verdict "Pass".\\
In a special case where the AI assistant claims to not have found any information, return the "Pass" verdict.\\
Your verdict must strictly be one of: "Fact-checked: [[True]]", "Fact checked: [[False]]", and "Fact checked: [[Pass]]"\\
\\
<|Start of factual info for \fontpurple{\{CHEMICAL\_NAME\}}|>\\
\fontpurple{\{CHEMICAL\_DOCUMENT\}}\\
<|End of factual info for \fontpurple{\{CHEMICAL\_NAME\}}|>\\

\midrule
\textbf{\textit{User Prompt}} \\
\midrule 
<|The Start of Assistant's Conversation with User|> \\
\#\#\# User:\\
\fontpurple{\{QUESTION\_1\}}\\
\#\#\# Assistant:\\
\fontpurple{\{ANSWER\_1\}}\\
<|The End of Assistant's Conversation with User|>\\
\bottomrule
\end{tabular}
}
\label{table:prompt-eval-db-consistency}
\end{table*}

\begin{table*}
\centering
\caption{LLM-as-a-judge on preference}
\resizebox{\textwidth}{!}{%
\begin{tabular}{l}
    \toprule
\textbf{\textit{System prompt}} \\
\midrule
Please act as an impartial judge and evaluate the quality of the responses provided by two AI assistants to the user question displayed below. \\
You should choose the assistant that follows the user’s instructions and answers the user’s question better. \\
Your evaluation should consider factors such as the helpfulness, relevance, accuracy, depth, 
and level of detail of their responses. \\
Begin your evaluation by comparing the two responses and provide a short explanation. \\
Avoid any position biases and ensure that the order in which the responses were presented does not influence your decision. \\
Do not allow the length of the responses to influence your evaluation. \\
Do not favor certain names of the assistants. \\
Be as objective as possible. \\
After providing your explanation, output your final verdict by strictly following this format: \\
"[[A]]" if assistant A is better, "[[B]]" if assistant B is better, and "[[C]]" for a tie. \\

\midrule
\textbf{\textit{User Prompt}} \\
\midrule 
<|The Start of Assistant A’s Conversation with User|>
\\ \#\#\# User:
\\ \fontpurple{\{QUESTION\_1\}}
\\\#\#\# Assistant A:
\\ \fontpurple{\{ANSWER\_1\_OF\_MODEL\_A\}}
\\\#\#\# User:
\\ \fontpurple{\{QUESTION\_2\_OF\_MODEL\_A\}}
\\\#\#\# Assistant A:
\\ \fontpurple{\{ANSWER\_2\_OF\_MODEL\_A\}}
\\<|The End of Assistant A’s Conversation with User|>
\\
\\<|The Start of Assistant B’s Conversation with User|>
\\\#\#\# User:
\\ \fontpurple{\{QUESTION 1\}}
\\\#\#\# Assistant B:
\\ \fontpurple{\{ANSWER\_1\_OF\_MODEL\_B\}}
\\\#\#\# User:
\\ \fontpurple{\{QUESTION\_2\_OF\_MODEL\_B\}}
\\\#\#\# Assistant B:
\\ \fontpurple{\{ANSWER\_2\_OF\_MODEL\_B\}}
\\<|The End of Assistant B’s Conversation with User|> \\
\bottomrule
\end{tabular}
}
\label{table:prompt-eval-preference}
\end{table*}

\clearpage
\onecolumn
\small
\begin{center}
\begin{longtable}{l}
\caption{English translation of the Hydroquinone document from Tox-Info Database.}
\label{tab:full-document-example}\\
\endfirsthead
\endhead
\textit{cont'd next page}
\endfoot
\endlastfoot

\toprule

\textbf{Abstract}\\
\midrule
Hydroquinone is a light tan, light gray, or colorless crystalline solid. It is primarily used in photographic developers,
\\ dye intermediates, polymerization inhibitors, and as a medication to treat skin hyperpigmentation. Acute toxicity from
\\ this substance appears to occur when high doses of hydroquinone are administered. Ingestion can cause tinnitus, 
\\ nausea, dizziness, a sense of suffocation, increased respiratory rate, vomiting, pallor, muscle twitching, headache,
\\ difficulty breathing, cyanosis, delirium, and collapse. In particular, long-term use on the skin at concentrations of 5\%
\\ or higher can cause pigmentation and depigmentation. The median lethal dose (LD₅₀) is 320 mg/kg for oral administration
\\ in rats, 115 mg/kg for intravenous administration in rats, and 182 mg/kg for subcutaneous administration in mice. 
\\ Carcinogenicity has been observed in laboratory animals, including an increased incidence of bladder cancer,
\\ hepatocellular neoplasms, and adenomas. In the IARC classification of carcinogens, hydroquinone is listed in Group 3 as
\\ "not classifiable as to its carcinogenicity to humans."
\\

\midrule

\textbf{1. General Substance Information}\\
\midrule
\textbf{English Name}: Hydroquinone
\\\textbf{Korean Name}: 히드로퀴논
\\\textbf{CAS No.}: 123-31-9
\\\textbf{Molecular Structure}:
\\\quad\includegraphics[width=3.0cm,height=!,valign=m]{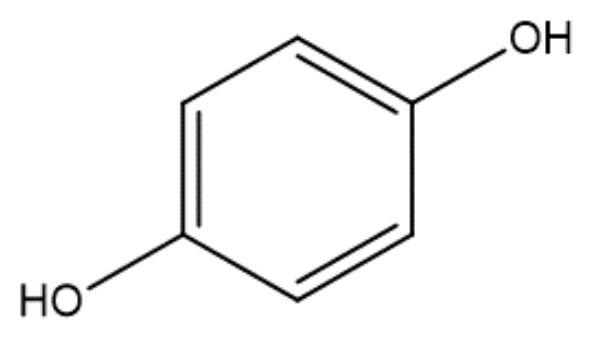}
\\\textbf{Molecular Formula}: C6-H6-O2 
\\\textbf{Molecular Weight}: 110.11
\\\textbf{Synonyms (English)}: 1,4-Dihydroxybenzene, Hydroquinol, Quinol, p-Dihydroxybenzene
\\\textbf{Synonyms (Korean)}: 히드로퀴논, 1,4-디히드록시벤젠, 히드로퀴놀, 퀴놀, p-디히드록시벤젠
\\\textbf{Color and Appearance}:
\\\quad 1. A white, crystalline solid.
\\\quad 2. Monoclinic prisms (sublimes); needle-like form from water; prismatic form from methanol.
\\\quad 3. Light tan, light gray, or colorless crystals.
\\\textbf{Odor}: Odorless
\\\textbf{Boiling Point}: 285-287\textdegree C
\\\textbf{Melting Point}: 170-171\textdegree C
\\\textbf{Vapor Pressure}: $1.9\times 10^{-5}$ mmHg at 25\textdegree C
\\\textbf{Density/Specific Gravity}: $1.330$ g/cm$^3$ at 20\textdegree C
\\\textbf{Solubility}:
\\\quad 1. Soluble in water at 72,000 mg/L at 25\textdegree C.
\\\quad 2. 7\% soluble in water at 25\textdegree C.
\\\quad 3. At 30\textdegree C, solubility per 100g of solvent is 46.4 g in ethanol, 28.4 g in acetone,
\\\quad 8.3 g in water, 0.06 g in benzene, and 0.01 g in carbon tetrachloride.
\\\quad 4. Freely soluble in ethyl ether.
\\\quad 5. Very soluble in carbon tetrachloride.
\\\textbf{GHS Pictogram}:
\\\quad\includegraphics[width=6.0cm,height=!,valign=m]{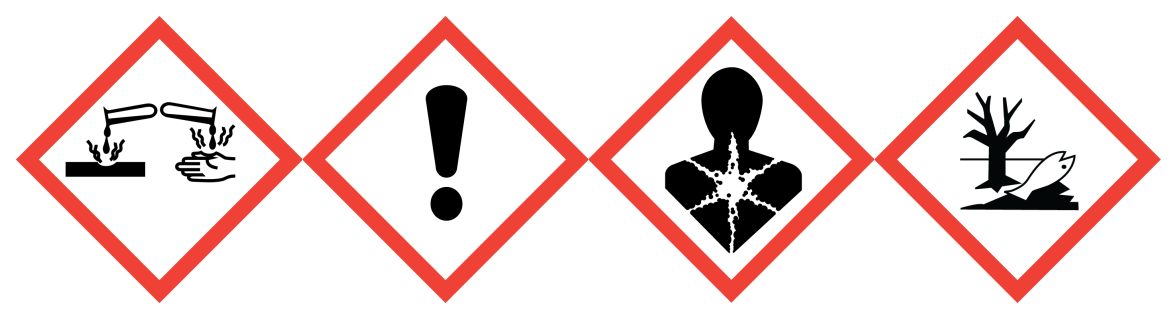}
\\

\midrule

\textbf{2. Usage}\\
\midrule
1. In photographic developers and the rubber industry as an antioxidant and antiozonant.
\\ 2. As a polymerization inhibitor to prevent polymerization by free radicals during the processing and storage of vinyl 
\\\quad monomers, and to stabilize unsaturated polyester resins.
\\ 3. In cosmetics (as an antioxidant, fragrance, and reducing agent) or as a polymerization inhibitor.
\\ 4. In topical preparations (as a skin bleaching and whitening agent).
\\ 5. As a reaction intermediate inhibitor used to stabilize monomers, paints, varnishes, and engine oils, and as a rust inhibitor 
\\\quad in cooling towers.
\\ 6. In photographic reducers and developers, as a reagent to detect small amounts of phosphate, and as an antioxidant.
\\ 7. In adhesives and hair dyes.
\\

\midrule
\pagebreak

\midrule
\textbf{3. Toxicity Information} - 3.1. Human Health Effects\\
\midrule
\textbf{Symptoms}:
\\\quad 1. Ingestion can cause tinnitus, nausea, dizziness, a sensation of choking, increased respiratory rate, vomiting, pallor, 
\\\quad\quad muscle cramps, headaches, difficulty breathing, cyanosis, delirium, and collapse (O'Neil, 2013).
\\\quad 2. The initial stimulatory effects on the central nervous system (CNS) from acute toxic doses of hydroquinone (HQ) are
\\\quad\quad well-known and similar to those of other phenolic resins. At low dose levels, clinical signs of salivation, tremors, and 
\\\quad\quad hypersensitivity appear. At lethal concentrations, CNS depression and respiratory issues following convulsions are 
\\\quad\quad commonly reported. Clinical signs typically occur shortly after oral and parenteral administration. Recovery is rapid 
\\\quad\quad and complete at sublethal doses (Bingham, E., 2001).
\\\quad 3. Reports on the effects of ingesting hydroquinone have shown clinical symptoms such as vomiting, abdominal pain, 
\\\quad\quad tachycardia, seizures, tremors, dyspnea (difficulty breathing), cyanosis, coma, loss of reflexes, and death 
\\\quad\quad (Sullivan, J.B., 1999).
\\
\\\textbf{Case Reports}:
\\\quad 1. A 43-year-old male, after 16 years of occupational exposure to hydroquinone in X-ray developer, was diagnosed
\\\quad\quad with myelodysplastic syndrome and acute myeloid leukemia. Cytogenetic studies showed chromosomal abnormalities
\\\quad\quad on chromosomes 5 and 7 (Regev L et al, 2012).
\\\quad 2. A gastrointestinal illness, characterized by acute symptoms of nausea, vomiting, abdominal pain, and diarrhea,
\\\quad\quad occurred in 544 crew members aboard a large U.S. Navy ship. This was found to have been caused by the
\\\quad\quad contamination of the cold-water system with hydroquinone from an onboard automatic photo-processing machine
\\\quad\quad (Hooper RR et al., 1978).
\\\quad 3. A fatality occurred due to the ingestion of a photographic developer containing hydroquinone. Hydroquinone 
\\\quad\quad was extracted from autopsy samples and confirmed by gas chromatography-mass spectrometry (GC-MS). The
\\\quad\quad concentrations of hydroquinone in urine, liver, and kidney were 3.4 $\mu$g/mL, 0.5 $\mu$g/g, and 0.2 $\mu$g/g,
\\\quad\quad respectively (Saito T et al, 1994).
\\
\\\textbf{Epidemiological Studies}:
\\\quad 1. The purpose of this study was to compare 33 workers exposed to a mixture of hydroquinone, trimethyl-hydroquinone,
\\\quad\quad and retinene-hydroquinone with 55 unexposed control subjects to determine the allergenic potential of the exposure.
\\\quad\quad The prevalence of respiratory symptoms was increased in the group exposed to the mixture. The exposed workers had 
\\\quad\quad a significantly higher incidence of coughing caused by smoke or cold air (P < 0.01). The prevalence of eczema in the
\\\quad\quad workplace was also high. Lung function values were significantly lower in the exposed group than in the control
\\\quad\quad group (P < 0.01). Workers exposed to the mixture had higher levels of IgG (P < 0.002) but not IgE than the control 
\\\quad\quad group (ESIS, 2009).
\\\quad 2. This study concerns the cohort mortality of 879 workers (22,895 person-years of follow-up) at a plant in Tennessee
\\\quad\quad (USA) where hydroquinone was manufactured and used for several decades. The average hydroquinone dust levels
\\\quad\quad ranged from 0.1 to 6.0 mg/cu m, with most of the plant's operational period at levels of 2 mg/cu m or higher.
\\\quad\quad The average employment duration was 13.7 years, with an average follow-up of 26.8 years after the initial exposure.
\\\quad\quad Relative risk estimates (standardized mortality ratios (SMRs)) for this cohort were derived by comparing them to the
\\\quad\quad general population of Tennessee and to a group unexposed to hydroquinone (from a plant of the same company in 
\\\quad\quad New York). The SMR for all causes of death (n=168) was significantly less than 1.0, as was the SMR for all cancers
\\\quad\quad (n=33). Only the colon (n=5) and lung (n=14) had more than three observed cases. Most site-specific SMRs were
\\\quad\quad well below 1.0. The results were similar for both comparison groups. Dose-response analyses for the selected cancer
\\\quad\quad sites showed no meaningful trend or heterogeneity (IARC, 1999).
\\\quad 3. A cohort incidence study was conducted among 837 Danish lithographers born between 1933 and 1942. In 1989,
\\\quad\quad a questionnaire was sent to the cohort participants to obtain information on occupational exposure, and responses
\\\quad\quad were received from 620 workers. Approximately one-quarter of the cohort participants reported regular exposure to
\\\quad\quad hydroquinone for photographic development. Relative risk estimates (standardized incidence ratios (SIRs)) for this
\\\quad\quad cohort were derived by comparing them to the general population of Denmark. A total of 24 cancers were
\\\quad\quad registered, and the SIR was 0.9. More than three cases only occurred in the skin, with no occurrences in other sites.
\\\quad\quad Five cases of malignant melanoma were observed, and 1.5 cases were predicted (SIR 3.4, 95\% confidence interval
\\\quad\quad 1.2-7.5). Two of these five cases were reported to have been exposed to hydroquinone (IARC, 1999).
\\
\\\textbf{Others}:
\\\quad 1. If the eyes are exposed to hydroquinone dust, it can lead to eye damage consisting of irritation, photosensitivity, 
\\\quad\quad tearing, corneal epithelial damage, and corneal ulcers (Sullivan, J.B., 1999).
\\

\midrule
\pagebreak
\midrule

\textbf{3. Toxicity Information} - 3.2. Animal Toxicity Test Information\\
\midrule
\textbf{Acute Toxicity}:
\\\quad 1. The absorption and metabolism of subcutaneously administered hydroquinone in Auratus goldfish occurred very 
\\\quad\quad rapidly in most tissues  and organs. While it showed no specific affinity for melanosomes, it did cause 
\\\quad\quad cytopathological changes in these pigment cells. Only melanin cells containing melanosomes that were present at
\\\quad\quad the time of treatment were destroyed (Chavin, 1971).

\\\quad 2. The oral LD50 values for hydroquinone in rats, mice, guinea pigs, cats, and dogs ranged from 70 to 550 g/kg, with
\\\quad\quad cats being the most sensitive. After a lethal dose was administered, excessive excitement, tremors, convulsions,
\\\quad\quad salivation, and vomiting were observed in cats within 90 minutes, and they died a few hours later (ACGIH, 2007).

\\\quad 3. An aqueous solution of hydroquinone was administered orally to male and female rats. Acute dermal toxicity was
\\\quad\quad also evaluated in rabbits. In the acute oral toxicity study, a single oral dose of 285, 315, 345, or 375 mg/kg was
\\\quad\quad given to five male and five female rats each. At all dose levels, the animals showed mild to moderate tremors and
\\\quad\quad mild convulsions within the first hour after administration. The acute oral LD50 value was >375 mg/kg. There were
\\\quad\quad no neurobehavioral effects or fatalities when 2,000 mg/kg of hydroquinone was applied to the skin of rabbits
\\\quad\quad under occlusion for 24 hours (Topping DC et al, 2007).

\\\quad 4. The effects of hydroquinone and its metabolite, 2,3,5-(tris-glutathione-S-yl)hydroquinone, on site-selective
\\\quad\quad cytotoxicity and cell proliferation in the rat kidney have been described. Male rats were treated with hydroquinone
\\\quad\quad (1.8, 4.5 mmol/kg, orally) or 2,3,5-(tris-glutathione-S-yl)hydroquinone (7.5 $\mu$mol/kg; 1.2–1.5 $\mu$mol/rat, intravenously),
\\\quad\quad  and blood urea nitrogen (BUN), urinary $\gamma$-glutamyltransferase (GGT), alkaline phosphatase (ALP), glutathione
\\\quad\quad S-transferase (GST), and glucose were measured as indicators of nephrotoxicity. In some rats, hydroquinone
\\\quad\quad (1.8 mmol/kg, oral) showed nephrotoxicity, and cell proliferation (BrDU incorporation) in the proximal tubule cells
\\\quad\quad near the S3M region was correlated with the degree of toxicity in each rat. At 4.5 mmol/kg, hydroquinone
\\\quad\quad significantly increased the urinary excretion of $\gamma$-GT, ALP, and GST. The hydroquinone metabolite,
\\\quad\quad 2,3,5-(tris-glutathione-S-yl)hydroquinone, caused an increase in BUN, urinary GGT, and ALP for up to 12 hours
\\\quad\quad after administration. In contrast, the maximum excretion of GST and glucose occurred after 24 hours. By 72 hours,
\\\quad\quad the concentrations of BUN and glucose recovered to control levels, while GGT, ALP, and GST remained slightly
\\\quad\quad elevated. Examination of kidney sections under a light microscope showed medullary necrosis in the S3M portion
\\\quad\quad of the proximal tubule (Peters MM et al, 1997).

\\
\\\textbf{Repeated Dose Toxicity}:
\\\quad 1. When rats were given hydroquinone at a concentration of 5\% (50,000 ppm) in their feed for 9 weeks, it caused
\\\quad\quad severe weight loss, aplastic anemia, bone marrow suppression, liver atrophy, and gastric mucosal ulcers and
\\\quad\quad hemorrhages (ACGIH, 2007).

\\\quad 2. Male and female F344 rats were topically treated with 0, 2.0, 3.5, and 5.0\% hydroquinone in an oil-in-water (ow)
\\\quad\quad emulsion cream for 13 weeks (5 days per week). Body weight, feed, and water intake were measured, and clinical
\\\quad\quad signs of toxicity and skin irritation were observed. Blood samples collected at the end of the study were analyzed
\\\quad\quad for hematological and clinical chemistry effects. Erythema was the only effect observed at the HQ cream application
\\\quad\quad site, and it subsided when the exposure was discontinued. Cell proliferation in the kidneys was evaluated using
\\\quad\quad bromodeoxyuridine (BrdU) labeling after 3, 6, and 13 weeks of treatment, but no changes indicating sustained cell
\\\quad\quad proliferation were seen. Renal histopathological lesions observed after oral HQ exposure in previous studies did not 
\\\quad\quad appear after dermal exposure. Therefore, topical exposure to HQ does not cause the nephrotoxicity observed in 
\\\quad\quad orally dosed F344 rats in previous studies (David RM et al, 1998).

\\\quad 3. In a 13-week oral toxicity study, four groups of 10 male and 10 female SD rats (approximately 7 weeks old at the 
\\\quad\quad start) were orally administered hydroquinone (0, 20, 64, and 200 mg/kg/day) 5 days per week. Brown urine was 
\\\quad\quad observed in both male and female rats at all doses. Males also had decreased food intake, which was only significant 
\\\quad\quad during the first week of the study (P < 0.05). There were no significant changes in body weight gain or food intake 
\\\quad\quad for females in any group throughout the study period. Signs of behavioral effects were observed at doses of 64 and 
\\\quad\quad 200 mg/kg. After the exposure period, the animals were sacrificed, and 6 males and females from each group were 
\\\quad\quad perfused for neuropathological examination. No treatment-related changes were observed in gross examination. In 
\\\quad\quad male rats, hydroquinone administration at 200 mg/kg resulted in decreased body weight gain, with a final body 
\\\quad\quad weight 7\% lower than the control group's average (WHO/IPCS, 1994).

\\\quad 4. Male and female rats were orally administered hydroquinone in feed at concentrations of 1, 1.25, 1.5, 2, and 4\% for
\\\quad\quad 90 days. Bone marrow atrophy/degeneration was observed in females at the 1\% concentration (corresponding to 
\\\quad\quad approximately 800 mg/kg bw/day). Body weight gain was reduced in both sexes starting at the 2\% concentration, 
\\\quad\quad and general symptoms were affected at the 4\% concentration (ESIS, 2009).

\\
\midrule
\pagebreak

\midrule
\textbf{3. Toxicity Information} - 3.2. Animal Toxicity Test Information \textit{(cont'd)}\\
\midrule
\textbf{Reproductive and Developmental Toxicity}:
\\\quad 1. Burnett (1976) reported the results of a teratology study on 12 hair dye complexes. The study groups were tested 
\\\quad\quad along with one positive and three negative control groups. Female SD rats (20/group) were topically administered 
\\\quad\quad 2 mL/kg (0.2\% hydroquinone) on gestational days 1, 4, 7, 10, 13, 16, and 19. No signs of toxicity were observed 
\\\quad\quad throughout the study. There were no differences between the control and hydroquinone-treated groups in the 
\\\quad\quad reported parameters (maternal toxicity, body weight and food intake, average number of corpora lutea, implantation 
\\\quad\quad sites, fetal absorption sites, average reabsorption rate per pregnancy, live fetuses, and sex ratio), and there were no 
\\\quad\quad significant changes in soft tissue or skeleton (WHO/IPCS, 1994).

\\\quad 2. Hydroquinone injected subcutaneously into male rats decreased fertility and prolonged the estrous cycle in females. 
\\\quad\quad However, this was not found in oral administration studies (dominant lethal studies and two-generation studies). In 
\\\quad\quad a developmental study in rats, a dose of 300 mg/kg bw was mildly toxic to the dams and reduced fetal body weight 
\\\quad\quad (WHO/IPCS, 1994).

\\\quad 3. A two-generation study in SD rats evaluated the effects of hydroquinone (HQ) on fertility and reproductive 
\\\quad\quad performance. HQ was administered orally at 0, 15, 50, and 150 mg/kg/day. F0 and F1 parental animals were dosed 
\\\quad\quad daily for at least 10 weeks before, during, and until the scheduled termination of cohabitation. At all dose levels, no 
\\\quad\quad adverse effects were observed on food consumption, survival, or reproductive parameters for the F0 and F1 parental 
\\\quad\quad animals. Mild, transient tremors were observed in F0 and F1 parental animals immediately after dosing at 
\\\quad\quad 150 mg/kg/day and in a single F0 male at 50 mg/kg/day. This tremor was intermittent and was thought to be due 
\\\quad\quad to the acute stimulatory effects of HQ on the nervous system. The body weights of F0 and F1 parental females were 
\\\quad\quad similar across all dose groups throughout the study. The body weights of F0 parental males were comparable to the 
\\\quad\quad control group throughout the study. For F1 males in the 50 and 150 mg/kg/day dose groups, statistically significant 
\\\quad\quad differences in body weight were noted during the pre-mating, cohabitation, and termination periods. No 
\\\quad\quad treatment-related effects on body weight, sex distribution, or survival were observed in the pups of either generation. 
\\\quad\quad Post-mortem examination revealed no treatment-related lesions in the F0, F1 parental animals, or pups. 
\\\quad\quad Histopathological examination of reproductive tissues and the pituitary from F0 and F1 parental animals at the high 
\\\quad\quad dose showed no changes related to HQ administration (Blacker AM et al, 1993).

\\
\\\textbf{Genotoxicity and Mutagenicity}:
\\\quad 1. A study was conducted to investigate whether hydroquinone induces aneuploidy in the bone marrow and 
\\\quad\quad embryonic cells of mice. Male (C57B1/CnexC3H/Cne) F1 mice were injected intraperitoneally with 0-400 mg/kg 
\\\quad\quad of hydroquinone. They were sacrificed at 6, 8, 18, and 24 hours post-injection, and their femurs and testes were 
\\\quad\quad removed. Bone marrow cells were analyzed for hyperdiploidy, ploidy, and micronucleated polychromatic 
\\\quad\quad erythrocytes. 80 mg/kg of hydroquinone was found to induce spermatid hyperdiploidy (Leopardi P et al, 1993).

\\\quad 2. Hydroquinone is generally negative in mutagenicity tests performed on several strains of S. typhimurium, with or 
\\\quad\quad without exogenous metabolic activation. When administered orally to adult male Drosophila via feed at 
\\\quad\quad 0.5-1.0 mg/mL, hydroquinone failed to induce sex-linked recessive lethal mutations and, in the mouse spot test, 
\\\quad\quad did not induce gene mutations in the somatic cells of mice (ACGIH, 2007).

\\\quad 3. Hydroquinone was not mutagenic in S. typhimurium strains TA98, TA100, TA1535, and TA1537, with or without 
\\\quad\quad exogenous metabolic activation. It did induce trifluorothymidine resistance in mouse L5178Y/TK lymphoma cells in 
\\\quad\quad the presence and absence of metabolic activation. A clear response was obtained in a test for the induction of 
\\\quad\quad sex-linked recessive lethal mutations in Drosophila administered hydroquinone via feed. Hydroquinone induced 
\\\quad\quad sister chromatid exchange in Chinese hamster ovary cells with or without exogenous metabolic activation and 
\\\quad\quad caused chromosomal aberrations in the presence of activation (NTP, 1989).

\\
\\\textbf{Eye/Skin Irritation}:
\\\quad 1. When a single lethal dose of hydroquinone was administered orally or subcutaneously to various laboratory animals,
\\\quad\quad non-specific effects on the nervous system such as hypersensitivity, tremors, and convulsions were observed.
\\\quad\quad Animals given sublethal oral doses recovered within a few days (WHO/IPCS, 1994).

\\\quad 2. Based on the absence of skin effects at the application site on rabbits, no signs of skin irritation were observed 
\\\quad\quad during skin toxicity studies at test doses similar to those used during the skin toxicity studies. According to OECD 
\\\quad\quad Guideline 404, more severe test conditions were applied in the skin irritation study than usual, with a 24-hour 
\\\quad\quad occlusive application (ECHA, 2020).

\\\quad 3. When an aqueous solution of hydroquinone at 5\% and 35\% concentrations was applied to the shaved dorsal 
\\\quad\quad skin of rats for 24 hours in an open application, no local effects on the skin were observed after 14 days of 
\\\quad\quad observation (ECHA, 2020).

\\
\midrule
\pagebreak

\midrule
\textbf{3. Toxicity Information} - 3.2. Animal Toxicity Test Information \textit{(cont'd)}\\
\midrule
\textbf{Immunotoxicity}:
\\\quad 1. The effect of hydroquinone on the production of calpain, an IL-1$\alpha$-processing enzyme by mouse bone marrow 
\\\quad\quad macrophages, was investigated. Bone marrow macrophages were harvested from the femurs of male mice and 
\\\quad\quad cultured for 4-6 hours with hydroquinone concentrations of 0, 1, 10, and 100 $\mu$M. After 4 hours of exposure to 
\\\quad\quad hydroquinone, the calpain-II content of the cytosolic and particulate fractions decreased by about 50\%. The level
\\\quad\quad  of calpain-I was unchanged. The effect of hydroquinone is specific to calpain-II in bone marrow macrophages, 
\\\quad\quad indicating a mechanism for the decrease of pro-IL-1$\alpha$ to IL-1$\alpha$ after exposure to benzene or hydroquinone 
\\\quad\quad (Miller ACK et al, 1994).

\\\quad 2. The effect of immunotoxic chemicals on in vitro proliferative responses was studied in human and rodent 
\\\quad\quad lymphocytes. Spleen cells from female B6C3F1 mice, F344 rats, and human peripheral blood lymphocytes were 
\\\quad\quad stimulated with a T-lymphocyte CD3 complex (anti-CD3 antibody), phytohemagglutinin, or several mitogens. 
\\\quad\quad They were incubated for 20 hours with hydroquinone at 0 to 10$^{-5}$ molar. The effect on cell proliferation was 
\\\quad\quad evaluated by measuring the uptake of tritiated thymidine. Dose-response curves were plotted, graphing the change 
\\\quad\quad in the degree of cell proliferation against the concentration of the control value. Hydroquinone caused a biphasic 
\\\quad\quad response in mouse and human lymphocytes: the response after stimulation was most pronounced in lymphocytes 
\\\quad\quad stimulated by the anti-CD3 antibody. In rat lymphocytes, only an inhibitory effect was observed 
\\\quad\quad (Lang DS et al, 1993).

\\\quad 3. Male Wistar rats were exposed to either a solvent or hydroquinone (HQ) intraperitoneally once a day every 2 days 
\\\quad\quad for a total of 22 days. The animals were sensitized to ovalbumin (OA) 10 days after exposure to the solvent or HQ, 
\\\quad\quad and aerosolized OA was inoculated 23 days later. HQ exposure did not change the number of circulating leukocytes 
\\\quad\quad but did impair the allergic inflammatory mechanism. A decrease in the contractility of ex vivo tracheal rings induced 
\\\quad\quad by OA in HQ-exposed animals and an impairment of mast cell degranulation in the mesentery after in situ OA 
\\\quad\quad inoculation were observed in the tissues. The specificity of the reduced response to OA was confirmed by the normal
\\\quad\quad tracheal contraction and mast cell degranulation in response to compound 48/80. The lower expression of 
\\\quad\quad co-stimulatory molecules CD6 and CD45R on OA-activated lymphocytes in HQ-exposed rats suggests that HQ 
\\\quad\quad exposure interferes with humoral signaling during allergic inflammation (Macedo SMD et al, 2007).

\\
\\\textbf{Others}:
\\\quad 1. In a 13-week neurotoxicity study, daily doses of 64 and 200 mg/kg of hydroquinone caused tremors, but no lasting
\\\quad\quad effects on behavior, motor activity, or neuropathology were observed. Tremors were not seen with a daily dose of 
\\\quad\quad 20 mg/kg for 13 weeks (Bingham, 2001).

\\\quad 2. An aqueous solution of hydroquinone (HQ) was administered orally to male and female SD rats. Sub-chronic 
\\\quad\quad exposure involved administering HQ aqueous solution at 0, 20, 64, and 200 mg/kg/day to study groups of 10 
\\\quad\quad animals per sex per group. A functional observational battery (FOB) was used to detect neurobehavioral effects at 
\\\quad\quad 1, 6, and 24 hours, and on days 7, 14, 30, 60, and 91 after HQ exposure. Daily clinical observations for each animal
\\\quad\quad were also recorded. Doses of 64 and 200 mg/kg of HQ produced noticeable behavioral effects, including acute 
\\\quad\quad tremors and decreased activity. The tremors occurred within one hour of dosing and disappeared by the six-hour 
\\\quad\quad mark. HQ administration did not change brain weights, but the average final body weight for males in the 
\\\quad\quad 200 mg/kg dose group was reduced by about 7\%. Neuropathological examination of the central and peripheral 
\\\quad\quad nervous systems, including specific modifications to myelin and axonal responses, did not reveal any lesions 
\\\quad\quad secondary to HQ administration or repeated CNS stimulation caused by HQ. Acute neurobehavioral effects 
\\\quad\quad indicating CNS stimulation were seen at oral doses of 64 mg/kg or higher. However, sub-chronic exposure at 
\\\quad\quad doses that caused repeated CNS stimulation did not lead to an aggravation of the acute stimulatory effects or to 
\\\quad\quad morphological changes in the central and peripheral nervous systems or nephrotoxicity over time 
\\\quad\quad (Topping DC et al, 2007).

\\\quad 3. In a 90-day study in rats using a functional observational battery, doses of 64 and 200 mg/kg of hydroquinone 
\\\quad\quad caused tremors, and the 200 mg/kg dose led to decreased general activity. Neuropathological examinations were 
\\\quad\quad negative (WHO/IPCS, 1994).

\\
\midrule
\textbf{3. Toxicity Information} - 3.3. Carcinogenicity\\
\midrule
\textbf{Carcinogenicity Classification}:
\\\quad IARC: 3 (Not Classifiable) / NTP: N/A (No data) / USEPA: N/A (No data)

\\

\\\textbf{Human Carcinogenicity Information}:
\\\quad 1. The evidence for hydroquinone's carcinogenicity in humans is inadequate. The evidence from experimental animals
\\\quad\quad for its carcinogenicity is limited. According to a comprehensive evaluation, hydroquinone cannot be classified as to 
\\\quad\quad its carcinogenicity to humans (Group 3) (IARC, 1999).

\\\quad 2. A3; Confirmed animal carcinogen with unknown relevance to humans (American Conference of Governmental 
\\\quad\quad Industrial Hygienists., 2014).

\\
\midrule
\pagebreak

\midrule
\textbf{3. Toxicity Information} - 3.3. Carcinogenicity \textit{(cont'd)}\\
\midrule
\textbf{Animal Carcinogenicity Test Information}:
\\\quad 1. Hydroquinone (HQ) is a "non-genotoxic" carcinogen that is generally negative in standard mutagenicity tests, and 
\\\quad\quad its mechanism of action is not fully known. HQ is metabolized into 2,3,5-tris(glutathione-S-yl)HQ (TGHQ), a highly 
\\\quad\quad toxic and redox-active compound. To confirm if TGHQ is a carcinogen in the kidneys, TGHQ was administered to 
\\\quad\quad Eker rats (2 months old) for 4 and 10 months. Eker rats are highly susceptible to kidney cancer development 
\\\quad\quad because they carry a germline mutation in the tuberous sclerosis 2 (Tsc-2) tumor suppressor gene. TGHQ-treated 
\\\quad\quad rats developed numerous toxic tubular dysplasias after just 4 months of treatment (2.5 $\mu$mol/kg, i.p.), which were 
\\\quad\quad rarely present in solvent-treated rats. These preneoplastic lesions indicate initial transformations within the tubules 
\\\quad\quad that undergo regeneration after damage by TGHQ, and adenomas developed within these lesions. After 10 months 
\\\quad\quad of treatment (2.5 $\mu$mol/kg for 4 months, 3.5 $\mu$mol/kg for 6 months), there was a 6-, 7-, and 10-fold increase in 
\\\quad\quad basophilic dysplasia, adenomas, and renal cell carcinomas, respectively. Most of these lesions were located in the 
\\\quad\quad outer stripe of the outer medulla, the same region affected by acute renal failure induced by TGHQ. Loss of 
\\\quad\quad heterozygosity (LOH) at the Tsc-2 locus in the toxic tubular dysplasias and tumors of TGHQ-treated rats was 
\\\quad\quad consistent with the loss of the Tsc-2 gene's tumor suppressor function by TGHQ. Therefore, although HQ is 
\\\quad\quad generally considered a non-genotoxic carcinogen, these results suggest that the formation of renal tumors by HQ 
\\\quad\quad is mediated by the formation of TGHQ, a nephrotoxic metabolite that induces a small but persistent regenerative 
\\\quad\quad proliferation and the loss of tumor suppressor gene function (Lau SS et al, 2001).

\\\quad 2. In a study to evaluate the carcinogenicity of hydroquinone in the mouse bladder, an unspecified number of mice 
\\\quad\quad were implanted with 10 mg cholesterol/20\% hydroquinone pellets (2 mg hydroquinone per mouse) and observed 
\\\quad\quad for 25 weeks. At 25 weeks, the incidence of bladder tumors in the surviving animals of the treated group (6 out of 
\\\quad\quad 19 mice) was significantly higher than the incidence in the group implanted with only cholesterol pellets (5 out of 
\\\quad\quad 57 mice) (ACGIH, 2007).

\\\quad 3. Toxicology and carcinogenicity studies were conducted in F344/N rats and B6C3F1 mice by administering 
\\\quad\quad hydroquinone (purity >99\%) orally for 14 days, 13 weeks, and 2 years. The 14-day study was performed by 
\\\quad\quad administering hydroquinone in corn oil at 63-1,000 mg/kg to rats and 31-500 mg/kg to mice, 5 days per week. 
\\\quad\quad In the 13-week study, doses for rats and mice ranged from 25-400 mg/kg. In the 14-day and 13-week studies, 
\\\quad\quad at doses that showed some signs of toxicity, the central nervous system, forestomach, and liver were identified as 
\\\quad\quad target organs in both species, and kidney toxicity was observed in rats. Based on these results, a 2-year study was 
\\\quad\quad conducted by orally administering 0, 25, and 50 mg/kg of hydroquinone in deionized water to 65 male and 
\\\quad\quad 65 female rats each (5 days/week). Each group of mice was administered 0, 50, and 100 mg/kg on the same 
\\\quad\quad schedule. After 15 months, 10 rats and 10 mice from each group were sacrificed and evaluated. The average body 
\\\quad\quad weights of male rats and mice at the high doses were about 5-14\% lower than the control groups' average body 
\\\quad\quad weights during the latter part of the study. There was no difference in survival rate between the treated and control 
\\\quad\quad groups of rats or mice. Nearly all male and female rats in all solvent-control and treated groups had nephritis, 
\\\quad\quad which was judged to be more severe in the high-dose male rats. Hyperplasia of the renal pelvis transitional 
\\\quad\quad epithelium and renal cortical cysts was increased in male rats. Renal tubular cell hyperplasia was observed in male 
\\\quad\quad rats given both high doses, and renal tubular adenomas were observed in 4/55 low-dose and 8/55 high-dose male 
\\\quad\quad rats. None were found in the solvent-control or female rats. Mononuclear cell leukemia in female rats showed an 
\\\quad\quad increased incidence in the treated groups (9/55 in the solvent control, 15/55 in the low dose, and 22/55 in the high 
\\\quad\quad dose). Compound-related lesions observed in the livers of high-dose male mice included heterotopic cysticosis, 
\\\quad\quad complex changes, and basophilic foci. The incidence of hepatocellular neoplasms, primarily adenomas, was 
\\\quad\quad increased in treated female mice (3/55; 16/55; 13/55). Follicular cell hyperplasia of the thyroid was increased in the 
\\\quad\quad treated mice (Kari FW et al, 1992).

\\
\midrule
\textbf{3. Toxicity Information} - 3.4. Toxicity Values\\
\midrule
\begin{tabular}{lllllll}
\midrule
\textit{Test Type} & \textit{Endpoint} & \textit{Species} & \textit{Sex} & \textit{Route} & \textit{Dose} & \textit{Cite}\\
\midrule
Human & DNEL & Human & - & Inhalation & = 1.05 mg/m³ & ECHA \\
Human & DNEL & Human & - & Dermal & = 1.66 mg/kg bw/day & ECHA \\
Human & DNEL & Human & - & Oral & = 0.6 mg/kg bw/day & ECHA \\
Acute & LD50 & Dog & - & Oral & = 299 mg/kg bw & European Commission, 2009 \\
Acute & LD50 & Cat & - & Oral & = 50 mg/kg bw & European Commission, 2009 \\
Acute & LD50 & Guinea Pig & - & Oral & = 550 mg/kg bw & European Commission, 2009 \\
Acute & LD50 & Guinea Pig & - & Dermal & > 1,000 mg/kg bw & European Commission, 2009 \\
Acute & LD50 & Rat & - & Oral & = 320 mg/kg & O'Neil, M.J. (ed.)., 2013 \\
Acute & LD50 & Rat & - & Intravenous & = 115 mg/kg & Lewis, R.J. Sr. (ed), 2004 \\
Acute & LD50 & Rat & - & Intraperitoneal & = 170 mg/kg & Lewis, R.J. Sr. (ed), 2004 \\
Acute & LD50 & Rat & - & Dermal & > 900 mg/kg bw & European Commission, 2009 \\
Acute & LD50 & Mouse & - & Subcutaneous & = 182 mg/kg & Lewis, R.J. Sr. (ed), 2004 \\
Acute & LD50 & Mouse & - & Oral & = 245 mg/kg & Lewis, R.J. Sr. (ed), 2004 \\
Acute & LD50 & Mouse & - & Intraperitoneal & = 100 mg/kg & Lewis, R.J. Sr. (ed), 2004 \\
Acute & LD50 & Rabbit & - & Intraperitoneal & = 125 mg/kg & Lewis, R.J. Sr. (ed), 2004 \\
Acute & LD50 & Rabbit & - & Oral & = 540 mg/kg bw & European Commission, 2009 \\
- & LDLo & Human & - & Oral & = 29 mg/kg & Deichmann, W.B., 1969 \\
- & TDLo & Human & - & Oral & = 170 mg/kg & Annales de M.L., 1927 \\
\midrule
\end{tabular}
\\
\midrule

\\
\midrule
\textbf{4. Toxicokinetics Information} - 4.1. Human Information\\
\midrule
1. When 2\% [14C]-hydroquinone was applied to the forearm of human subjects (n=4) in an unspecified cream, 
\\\quad the hydroquinone moved rapidly and consistently into the stratum corneum, and the radioactive label was detected 
\\\quad in plasma samples within 0.5 hours. During 8 hours of plasma sampling, hydroquinone concentration peaked at 
\\\quad 4 hours (0.04 equivalents/mL). After a 24-hour application of a 2\% cream to the forehead of 6 male volunteers, 
\\\quad the recovery of hydroquinone in the urine was 45.3\% (SD=11.2\%) (DHHS/NTP, 2009).

\\ 2. The dermal absorption of hydroquinone in humans is less efficient than oral administration. After a 24-hour 
\\\quad application of hydroquinone (2.0\% in alcohol) to the forehead of volunteers (6 males per formulation), the average 
\\\quad percutaneous absorption, measured as the elimination of hydroquinone via urine, was 57\% (SD=11\%). Peak 
\\\quad elimination occurred within 12 hours and was completely eliminated by 5 days. The addition of a sunscreen 
\\\quad (3.0\% Escalol 507) significantly reduced absorption (26\%, SD=14\%), while the addition of a penetration enhancer 
\\\quad (0\% Azone) did not significantly increase absorption, regardless of the presence of sunscreen (35\%, SD 17, and 
\\\quad 66\%, SD 13\%, respectively) (DHHS/NTP, 2009).

\\ 3. Oral administration of hydroquinone results in a high absorption rate. After consuming food containing 
\\\quad hydroquinone, peak plasma hydroquinone concentrations (5-fold increase) and maximum hydroquinone excretion 
\\\quad (12-fold increase) were reported in humans 2-3 hours later (DHHS/NTP, 2009).

\\
\midrule
\textbf{4. Toxicokinetics Information} - 4.2. Animal Information\\
\midrule
\textbf{Absorption}:
\\\quad 1. A toxicological review of hydroquinone includes several reports that hydroquinone is absorbed relatively quickly 
\\\quad\quad through oral administration, including a study on rats that ingested 3\% hydroquinone in a photographic developer. 
\\\quad\quad In addition, in SD and F344 rats administered 350 mg/kg, over 90\% was measured to be absorbed into the blood, 
\\\quad\quad with peak concentration observed within 1 hour (DHHS/NTP, 2009).

\\\quad 2. The percutaneous absorption rate of hydroquinone (HQ) through human stratum corneum and full-thickness rat 
\\\quad\quad skin was measured in vitro using a 5\% aqueous HQ solution as the donor solution. The study was conducted using 
\\\quad\quad an infinite dose of an aqueous solution containing [14C]-labeled HQ in a Franz-type diffusion cell. The measured 
\\\quad\quad absorption rate of HQ (mean ± SD) through human stratum corneum was 0.52 ± 0.13 $\mu$g/cm²/hour, and for rat 
\\\quad\quad skin, it was 1.1 ± 0.65 $\mu$g/cm²/hour. The ratio of the permeability coefficient (Kp) (rat/human) was 2.4. HQ would 
\\\quad\quad be classified as slow in terms of its absorption through the human stratum corneum (Barber ED et al, 1995).

\\
\\\textbf{Metabolism}:
\\\quad 1. This study investigated the metabolism of hydroquinone in male SD rats with and without prior hydroquinone 
\\\quad\quad treatment. A single oral dose of [14C]-hydroquinone was administered at concentrations of 5, 30, and 200 mg/kg. 
\\\quad\quad In one study, male rats were given 200 mg/kg of hydroquinone orally for 4 consecutive days, followed by a single 
\\\quad\quad dose of 200 mg/kg of [14C]-hydroquinone. In a separate study, rats were given a diet containing 5.6\% unlabeled 
\\\quad\quad hydroquinone for 2 days or a single oral dose of 311 mg/kg of [14C]-hydroquinone. The excretion pattern of 
\\\quad\quad [14C]-hydroquinone and its metabolites was similar in both single- and repeat-dose rats. Following a single dose of 
\\\quad\quad 200 mg/kg of [14C]-hydroquinone, 91.9\% was excreted in the urine within 2-4 days. 3.8\% was excreted in the 
\\\quad\quad feces, 0.4\% in the breath, and 1.2\% remained in the carcass. The radioactivity was distributed throughout all tissues, 
\\\quad\quad with higher concentrations in the liver and kidneys. The concentration of [14C] in the tissues decreased between 
\\\quad\quad 48 and 96 hours. The radiolabeled compounds in the urine were hydroquinone (1.1-8.6\% of dose), hydroquinone 
\\\quad\quad monosulfate (25-42\%), and hydroquinone monoglucuronide (56-66\%). Similar results were observed for rats given 
\\\quad\quad hydroquinone in their diet. There was no significant increase in absolute or relative liver weight, liver microsomal 
\\\quad\quad protein concentration, or cyb-5, cyP450, and cyC reductase activity in rats that received repeated doses of 
\\\quad\quad 200 mg/kg of hydroquinone. The cyP450 value was slightly but significantly decreased in rats receiving repeated 
\\\quad\quad hydroquinone doses (Divincenzo GD et al, 1984).

\\\quad 2. The metabolite 2-(S-glutathionyl)hydroquinone is formed when glutathione is added to microsomal incubation 
\\\quad\quad mixtures containing benzene or phenol. This metabolite is formed by the conjugation of benzoquinone, an oxidation
\\\quad\quad product of hydroquinone. However, the glutathione conjugate or its mercapturate, 
\\\quad\quad N-acetyl-S-(2,5-dihydroxyphenyl)-L-cysteine, has not been identified as a metabolite from the in vivo metabolism of 
\\\quad\quad benzene, phenol, or hydroquinone. To determine if the hydroxylated mercapturate is produced in vivo, male SD rats 
\\\quad\quad were treated with benzene (600 mg/kg), phenol (75 mg/kg), or hydroquinone (75 mg/kg), and urine was collected 
\\\quad\quad for 24 hours. Through HPLC coupled with an electrochemical detector (ECD), the presence of the metabolite 
\\\quad\quad N-acetyl-S-(2,5-dihydroxyphenyl)-L-cysteine was confirmed chromatographically and electrochemically. 
\\\quad\quad The metabolite was isolated from urine samples and treated with diazomethane to form the methyl ester derivative 
\\\quad\quad of N-acetyl-S-(2,5-dimethoxyphenyl)-L-cysteine. The mass spectrum obtained from this sample was identical to 
\\\quad\quad that of the standard derivative. These results indicate that benzene, phenol, and hydroquinone are metabolized to 
\\\quad\quad benzoquinone in vivo and excreted as the mercapturate, N-acetyl-S-(2,5-dihydroxyphenyl)-L-cysteine 
\\\quad\quad (Nerland, 1990).

\\\quad 3. When hydroquinone (HQ) was orally administered to male and female rats, less than 3\% of the excreted amount 
\\\quad\quad was the parent compound, indicating extensive metabolism. The major urinary metabolites of HQ were glucuronide 
\\\quad\quad and O-sulfate conjugates, which accounted for 45-53\% and 19-33\% of the oral dose, respectively. Less than 5\% of 
\\\quad\quad the metabolites were identified as mercapturate conjugates of HQ (English, 2005).

\\
\midrule
\pagebreak

\midrule
\textbf{4. Toxicokinetics Information} - 4.2. Animal Information \textit{(cont'd)}\\
\midrule
\textbf{Distribution}:
\\\quad 1. When a single intravenous dose of radiolabeled hydroquinone was administered to rats at 1.2–12 mg/kg, 
\\\quad\quad radioactivity (hydroquinone or its metabolites) was detected in the bone marrow and thymus within 2 hours.
\\\quad\quad Radioactivity was also detected in the liver and bone marrow of these rats for up to 24 hours. Regardless of whether 
\\\quad\quad a single or repeated oral dose was given, radioactivity was found in various tissues of rats, with the highest 
\\\quad\quad concentrations observed in the liver and kidneys. Following intravenous administration of radiolabeled hydroquinone 
\\\quad\quad to dogs, radioactivity was found in the skin, liver, and intestines. When radiolabeled hydroquinone was administered 
\\\quad\quad intraperitoneally to mice at 75 mg/kg, radioactivity was detected as being covalently bound to proteins in the liver, 
\\\quad\quad kidneys, blood, and bone marrow, with the specific activity in the liver being 10 times higher than that in the bone 
\\\quad\quad marrow (DHHS/NTP, 2009).

\\
\\\textbf{Excretion}:
\\\quad 1. In rabbits, less than 1\% of the administered dose was excreted unchanged, and about 80\% of the dose was 
\\\quad\quad recovered as conjugates in the urine (ACGIH, 2007).

\\\quad 2. After a single administration of radiolabeled hydroquinone (200 mg/kg) to rats, mass analysis showed that about 
\\\quad\quad 90\% was excreted in the urine and about 4\% in the feces within 48 hours. 1.2\% remained in the carcass, and 0.4\% 
\\\quad\quad was exhaled through respiration (ACGIH, 2007).

\\\quad 3. When hydroquinone was administered orally to F344 rats, peak blood concentrations were observed within 
\\\quad\quad 20 minutes, and 87-94\% was eliminated in the urine and cage rinse, while 1-3\% was eliminated in the feces 
\\\quad\quad up to 48 hours later (DHHS/NTP, 2009).

\\
\midrule
\textbf{5. Emergency Treatment Information} - 5.1. General Treatment\\
\midrule
\textbf{Inhalation Exposure}:
\\\quad Breathe fresh air and rest. Seek medical attention if necessary.

\\\textbf{Skin Exposure}:
\\\quad Wash contaminated clothing with plenty of water. Remove contaminated clothing and wash the skin with plenty of water.

\\\textbf{Eye Exposure}:
\\\quad First, wash the eyes with plenty of water for several minutes. If possible, remove contact lenses and seek medical attention.

\\\textbf{Oral Exposure}:
\\\quad Rinse the mouth. Seek medical attention.

\\
\midrule
\textbf{5. Emergency Treatment Information} - 5.2. Specific Treatment\\
\midrule
1. Immediate First Aid: Ensure proper decontamination has been performed. If the patient isn't breathing, begin 
\\\quad artificial respiration. It's best to use a demand valve, bag-valve mask, or pocket mask if you're trained. Perform 
\\\quad cardiopulmonary resuscitation (CPR) if necessary. Immediately flush contaminated eyes with a continuous stream of 
\\\quad water. Do not induce vomiting. If vomiting occurs, keep the airway open and position the patient leaning forward or 
\\\quad on their left side (with the head lower if possible) to prevent aspiration. Keep the patient comfortable and maintain 
\\\quad a normal body temperature. Seek medical attention (Currance, P.L., 2007).

\\ 2. Basic Treatment: Establish an open airway (oropharyngeal or nasopharyngeal airway if needed). Suction if necessary, 
\\\quad monitor for signs of respiratory insufficiency, and assist with artificial respiration if required. Administer oxygen at 
\\\quad 10–15 L/min via a non-rebreather mask. Monitor for and treat pulmonary edema if necessary. Monitor for and treat 
\\\quad shock if necessary. Anticipate seizures and treat if necessary. If a substance has entered the eyes, immediately flush 
\\\quad them with water. Continuously irrigate each eye with 0.9
\\\quad ingested, rinse the mouth. Administer 5 mL/kg of water up to 200 mL if the patient can swallow, has a strong gag 
\\\quad reflex, and isn't drooling. Administer activated charcoal (Currance, P.L., 2007).

\\ 3. Advanced Treatment: Consider orotracheal or nasotracheal intubation to control the airway for unconscious patients, 
\\\quad those with severe pulmonary edema, or those experiencing severe respiratory distress. Consider drug therapy for 
\\\quad pulmonary edema. Positive-pressure ventilation with a bag-valve mask may be helpful. Consider drug therapy for 
\\\quad pulmonary edema. For severe bronchospasm, consider administering a beta-agonist like albuterol. Initiate 
\\\quad intravenous administration of D5W for heart rhythm monitoring and arrhythmia treatment if needed. SRP: 
\\\quad Maintain an open IV, minimum flow rate. If there are signs of hypovolemia, use 0.9
\\\quad Ringer's solution. For hypotension with signs of hypoglycemia, administer fluids cautiously. Consider using fluids 
\\\quad and vasopressors to manage hypotension. Watch for signs of fluid overload. If the patient has severe hypoxia, 
\\\quad cyanosis, and symptoms of cardiac damage that do not respond to oxygen, administer a 1
\\\quad solution. Treat seizures with diazepam or lorazepam. Use proparacaine hydrochloride to aid eye irrigation 
\\\quad (Currance, P.L., 2007).

\\
\bottomrule
\end{longtable}
\end{center}
\twocolumn

\end{document}